\documentclass[10pt,twocolumn,letterpaper]{article}
\usepackage[accsupp]{axessibility}
\usepackage{cvpr}              

\usepackage{graphicx}
\usepackage{amsmath}
\usepackage{amssymb}
\usepackage{booktabs}
\usepackage{wrapfig}
\usepackage{float}
\usepackage{multirow}
\usepackage{amsfonts}
\usepackage{bbm}
\usepackage{accents}
\usepackage{derivative}
\usepackage[pagebackref,breaklinks,colorlinks]{hyperref}

\makeatletter
\newcommand*{\rightharpoonupfill@}{\arrowfill@\relbar\relbar\rightharpoonup}
\newcommand*{\overrightharpoonup}{\mathpalette{\overarrow@\rightharpoonupfill@}}

\makeatother

\usepackage[capitalize]{cleveref}
\crefname{section}{Sec.}{Secs.}
\Crefname{section}{Section}{Sections}
\Crefname{table}{Table}{Tables}
\crefname{table}{Tab.}{Tabs.}

\def\cvprPaperID{3} 
\def\confName{CVPR}
\def\confYear{2022}

\begin{document}

\title{Reconstruct from BEV: A 3D Lane Detection Approach based on Geometry Structure Prior}

\author{
   {Chenguang Li}\thanks{Both authors contributed equally.}${~~^{1}}$,
   {Jia Shi}\footnotemark[1]
   \thanks{This work was done during internship at SenseTime Research.}${~~^{1,2}}$,
   {Ya Wang}\footnotemark[2]${~~^{1,3}}$,
   {Guangliang Cheng}\thanks{Guangliang Cheng is the corresponding author.}${^{~~1,4}}$ \\
   ${^1}${SenseTime Research} \quad
   ${^2}${Robotics Institute, Carnegie Mellon University} \\
   ${^3}${University of Tuebingen} \quad
   ${^4}${Shanghai AI Laboratory} \\
   {\tt\small \{lichenguang, wangya\}@senseauto.com \, jiashi@andrew.cmu.edu \, guangliangcheng2014@gmail.com}
}

\maketitle
\begin{abstract}

In this paper, we propose an advanced approach in targeting the problem of monocular 3D lane detection by leveraging geometry structure underneath the process of 2D to 3D lane reconstruction. Inspired by previous methods, we first analyze the geometry heuristic between the 3D lane and its 2D representation on the ground and propose to impose explicit supervision based on the structure prior, which makes it achievable to build inter-lane and intra-lane relationships to facilitate the reconstruction of 3D lanes from local to global. Second, to reduce the structure loss in 2D lane representation, we directly extract BEV lane information from front view images, which tremendously eases the confusion of distant lane features in previous methods. Furthermore, we propose a novel task-specific data augmentation method by synthesizing new training data for both segmentation and reconstruction tasks in our pipeline, to counter the imbalanced data distribution of camera pose and ground slope to improve generalization on unseen data. Our work marks the first attempt to employ the geometry prior information into DNN-based 3D lane detection and makes it achievable for detecting lanes in an extra-long distance, doubling the original detection range. The proposed method can be smoothly adopted by other frameworks without extra costs. Experimental results show that our work outperforms state-of-the-art approaches by 3.8\% F-Score on Apollo 3D synthetic dataset at real-time speed of 82 FPS without introducing extra parameters.

\end{abstract}

\begin{figure}
    \centering
    \includegraphics[width=0.95\linewidth]{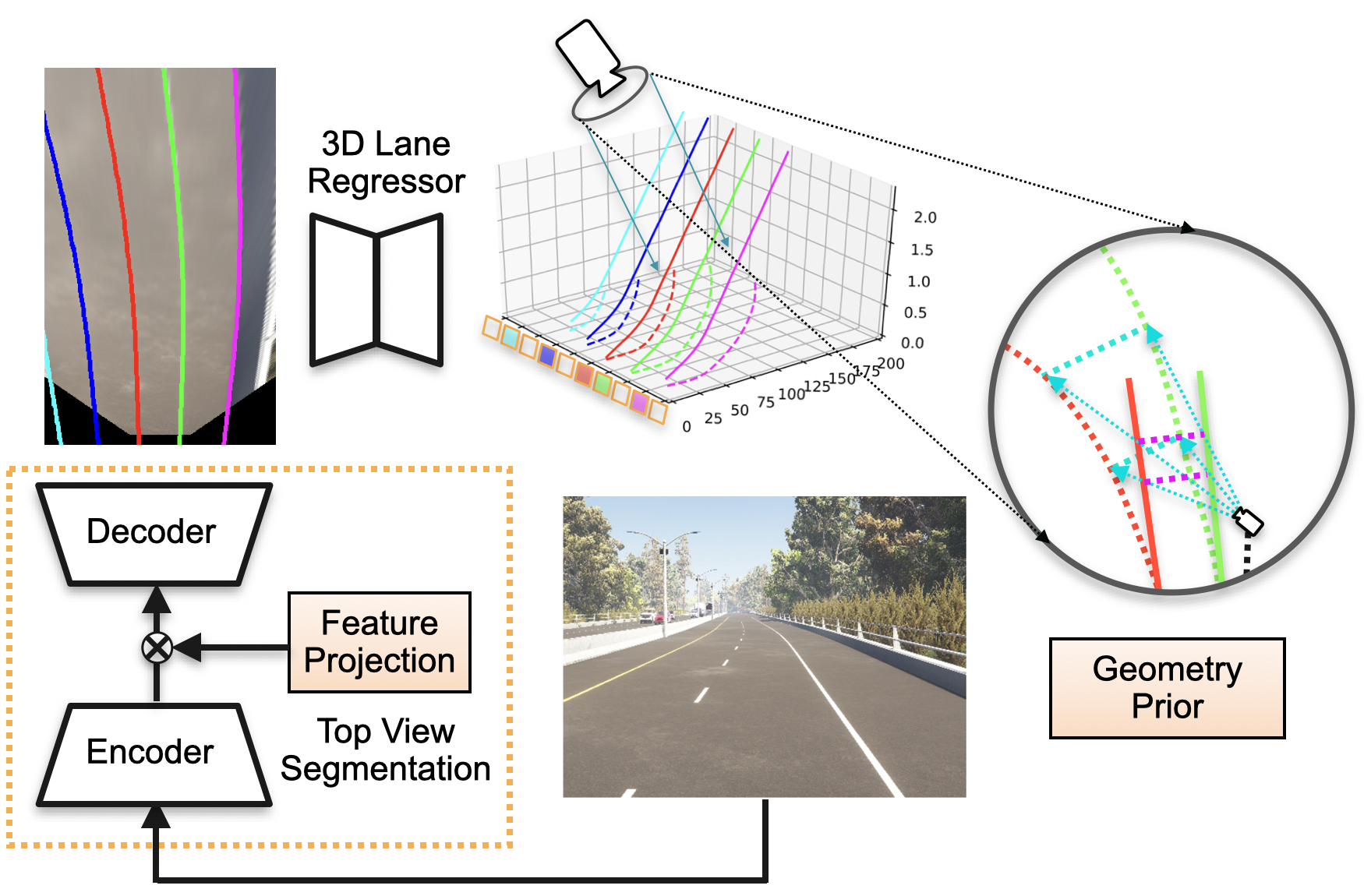}
    \caption{\textbf{Framework for 3D lane detection.} The framework is composed of a BEV (top view) segmentation network and an anchor-based 3D lane regressor. 
    The features extracted from a front view image are projected onto BEV to generate the segmentation mask directly under BEV supervision. The anchor representation of 3D lanes is then estimated from BEV lane mask. Moreover, we explicitly utilize the geometry prior from 3D to 2D projection for 3D lane reconstruction.}
    \label{fig:main_plot}
\vspace{-0.5cm}
\end{figure}

\section{Introduction}
\label{sec:intro}
\footnote{Proceedings of the CVPR 2022 Workshop of Autonomous Driving}Lane detection is a fundamental and challenging task for autonomous vehicles. Recently, numerous researches \cite{butakov2014personalized, chen2017end, yu2020bdd100k , pan2018spatial, homayounfar2019dagmapper} are conducted in this field for the extensive landing of high-level driving automation.
Accurate lane detection will provide essential signals for the localization \& mapping, decision making and path planning in autonomous driving systems (ADS), while most of the existing methods are built on the assumption of a flat ground plane \cite{bertozzi1998gold, jiang2010computer}, which is not necessarily satisfied in practice.
For instance, to restore the lane information, inverse perspective transformation (IPM) is often applied for projecting lanes from front view to Bird's-Eye-View (BEV or top view) \cite{mallot1991inverse, pomerleau1995ralph, aly2008real, kim2008robust}. However, the homography matrix for IPM is usually calibrated by point pairs on the flat ground \cite{neven2018towards}. When encountering roads with non-zero slope  like uphill or downhill situations, lane detection on 2D planar geometry would not accurately reflect the actual structure in real world, thus the ADS may malfunction due to the unexpected coordinate mapping. To solve this, it is critical to restore the height information by detecting lanes in 3D space. With accurate 3D lane estimation, most applications like lane keeping assistant (LKA) and lane departure warning (LDW) would be more reliable and robust, especially in non-flat ground scenarios.

In order to bridge the discrepancy between 2D and 3D lane detection, methods like \cite{dickmanns1992recursive, xiong20183d, coulombeau2002vehicle, nedevschi20043d, bai2018deep, dementhon1987zero} were proposed. With the rapid development of the Convolutional Neural Network, deep learning based approaches \cite{garnett20193d, guo2020gen, efrat20203d, jin2021robust} have made tremendous progress in resolving this problem. Although the existing DNN-based methods have achieved promising results, they attempt to approach 3D lane detection as a point-wise regression task while the underlying geometry structure of 3D lanes seems to be ignored, which may result in unstable prediction with inaccurate structure.

In this paper, we consider 3D lane detection as a reconstruction problem from the 2D image to the 3D space. We propose that the geometry prior of 3D lanes should be explicitly imposed during the training process for fully utilizing the structural constraints of the inter-lane and intra-lane relationship, and the height information of 3D lanes can be extracted from the 2D lane representation.
We first analyze the geometry relationship between 3D lane and its 2D representation, and propose an auxiliary loss function based on the geometry structure prior. We also demonstrate that the explicit geometry supervision would boost noise elimination, outlier rejection, and structure preservation for 3D lanes.
\par
\begin{figure*}[ht]
    \centering
    \vspace{-0.4cm}
    \includegraphics[width=0.65\linewidth]{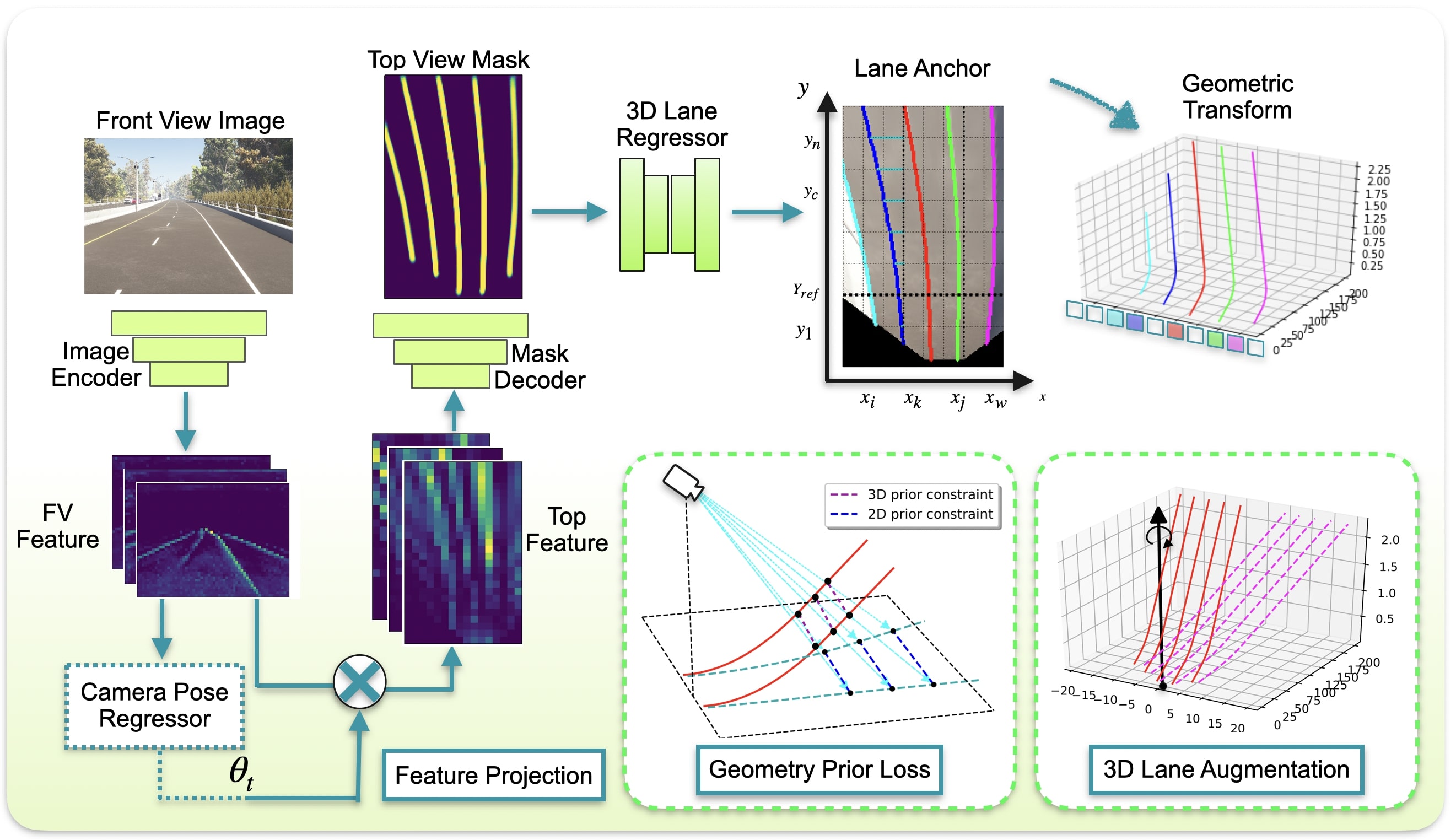}
    \caption{\textbf{Overall pipeline of the proposed method.}}
    \label{fig:flowchart}
\vspace{-0.4cm}
\end{figure*}

 Second, in order to reduce the structural information loss in 2D plane, we redefine the pipeline by conducting lane segmentation with BEV supervision instead of front view supervision, which addresses the issue of feature confusion due to the perspective distortion \cite{nieto2008robust} on the far side.
Lastly, we propose a novel task-specific data augmentation method on 3D lanes, which synthesizes new data by applying pitch, roll and yaw rotation on the original data.
This augmentation could generate new data with various 3D ground plane steepness and road structure patterns, which eases the imbalanced distribution of ground plane slope and camera pose. Figure~\ref{fig:main_plot} shows the proposed framework, and a more detailed pipeline is shown in Figure~\ref{fig:flowchart}. 

In general, our contributions can be summarized as follows: 1) A novel loss function that enables explicit supervision based on the geometry structure prior of lanes in 3D space for stable reconstruction from local to global. 2) A 2D lane feature extraction module with direct supervision from BEV for maximum retention of lane structural information especially in the far end. 3) A task-specific data augmentation method for 3D lane detection aiming to counter the imbalanced data distribution of ground slope and camera pose to improve generalization on rare cases.

\section{Related Work}

There are several conventional methods \cite{dickmanns1992recursive, xiong20183d, coulombeau2002vehicle, nedevschi20043d, bai2018deep, dementhon1987zero} that trickle the problem of lane detection in the 3D space. 
Dickmanns \textit{et al.}~\cite{dickmanns1992recursive} utilize a multi-sensor approach with temporal filtering to estimate the vertical curvature of 3D lane. Xiong \textit{et al.}~\cite{xiong20183d} adopt a B-spline curve to model lanes in 3D space, and resolve the depth ambiguity with known constant lane width prior. Coulombeau \textit{et al.}~\cite{coulombeau2002vehicle} model 3D road with camera pose and road curvature to estimate 3D lane shape from road points with optimization algorithm. Nedevschi \textit{et al.}~\cite{nedevschi20043d} apply stereo-vision for estimating 3D road surface iteratively with tracking process. Bai \textit{et al.}~\cite{bai2018deep} utilize multi-view sensor approach by combining LiDAR and camera for detecting lanes in real world. Dementhon \textit{et al.}~\cite{dementhon1987zero} model roads as a Brooks ribbon in space with constant length assumption and iteratively construct the lane plane with horizontal lane width generator. However, most of these geometry based methods build up parametric models for lanes which are not robust enough for complex roads and would fall short on detection range.

Recently, DNN-based methods \cite{garnett20193d, guo2020gen, efrat20203d, jin2021robust} have shown impressive results on monocular 3D lane detection. Garnett \textit{et al.} propose 3D-LaneNet \cite{garnett20193d}, an end to end dual-pathway \cite{he2016accurate} network bridging IPM features and 3D lane regression. The image-view pathway extracts and preserves dense features from images to estimate the camera pose, and the top-view pathway then predicts the 3D lanes from BEV features projected from the image-view pathway. Furthermore, as an analogy to single-shot detection \cite{redmon2016you, liu2016ssd}, a column-based anchor representation on the ground plane is proposed to encode each 3D lane instance.

On the top of 3D-LaneNet \cite{garnett20193d}, following recent anchor-free methods \cite{duan2019centernet, tian2019fcos}, Efrat \textit{et al.}\cite{efrat20203d} propose a more generalized semi-local tiled anchor representation to replace the previous column-based anchors, which divides continuous lane markings into short segments and encapsulates each of them into a non-overlapping cell. This novel anchor representation enables the generalization of complex topology like split, merge and horizontal lanes.

Gen-LaneNet\cite{guo2020gen} decouples 3D lane extraction from the image features by proposing a two-stage pipeline, which allows 3D lane estimation to be only dependent on 2D lane masks rather than the original images. This pipeline makes it possible to reduce the demand for expansive 3D lane labeling and utilize large 2D lane detection datasets to train a more robust lane feature extractor. Gen-LaneNet\cite{guo2020gen} also set up the anchor under virtual BEV projection to refine the feature misalignment in 3D-LaneNet\cite{garnett20193d}.

Besides, Jin \textit{et al.}\cite{jin2021robust} utilize a dual attention module \cite{fu2019dual} with the framework of Gen-LaneNet\cite{guo2020gen} to capture lane-to-lane and pixel-wise attention. Also, a linear interpolation loss is proposed to sample lane points within each anchor more densely. This method improves the accuracy yet introduces extra parameters and computational cost.

The up-to-date DNN-based methods are able to show promising results in monocular 3D lane detection. However, most of the existing methods seem to treat this problem only as a local lane regression task with 2D anchors, lacking explicit guidance on reconstructing the height of lanes from the 2D representation. 3D-LaneNet~\cite{garnett20193d} tries to bridge image feature encoding directly with lane decoding with a heavy VGG\cite{simonyan2014very} encoder, without deeper analysis of how the 3D lanes are detected from the image encoding. Gen-LaneNet~\cite{guo2020gen} points out that the estimation of 3D lane height is equivalent to estimating the vector field moving nonparallel lane points on the virtual BEV to corresponding 3D positions.
This heuristic releases the demand of utilizing a heavy network to encode image features in previous work. However, no further quantitative analysis or explicit supervision is applied from this perspective. Similarly, 3D-LaneNet+~\cite{efrat20203d} proposes a novel semi-local cell-based anchor, but this only enriches the structure representation on 2D space, rather than targeting the recovery of 3D lane information directly.

Monocular 3D lane detection relies on the accurate reconstruction of the 3D lane structure, in which the height information can be decoded from the 2D features.
Previous DNN-based methods \cite{garnett20193d, guo2020gen, efrat20203d, jin2021robust} choose to apply supervision on 2D anchors and a height dimension independently, however the underlying geometry structure information of lanes in the 3D space seems not to be fully utilized, which makes it challenging to accurately recover the 3D lanes. We propose that the structure of 3D lanes and the 2D-3D relationship following geometry prior should be jointly optimized in the whole process of 3D lane detection.

\section{Method}
\label{sec:geoprior}
The proposed method takes a single RGB image from the front-view camera, and outputs a group of lane instances in the 3D world space. Following the basic assumption in the previous literature\cite{garnett20193d, guo2020gen} and existing dataset~\cite{guo2020gen}, we assume that the camera is installed with zero roll or yaw respect to the world coordinate, and only has pitch variance due to vehicle fluctuation. We establish the world coordinate as the ego-vehicle coordinate with starting points as the perpendicular projection of camera center on the road. Figure~\ref{fig:lane_width} shows the world coordinate center at point $O$, and camera center at point $C$, with camera pitch $\theta$.

\subsection{Geometry in 3D Lane Detection}

In most cases, lanes in 3D space are formed by a group of smooth parallel curves on the 3D road surface\cite{dementhon1987zero}. Instead of simply predicting discrete points of lane markings, an accurate reconstruction of 3D lanes should include the re-establishment of the reasonable geometry structure in 3D space.
However, most of the existing DNN-based methods choose to achieve 3D lane detection in a data-driven manner with only point-wise supervision, which may not result in robust preservation of 3D lane geometry and would be vulnerable to outliers under extreme lane structures because of the absence of structural guidance.
As a result, the geometry prior should be utilized to explicitly guide the learning of 3D lanes. We will first review the view projection systems proposed in previous literature and then analyze the structure prior under the existing projection system. 

\begin{figure}[t]
    \centering
    \includegraphics[width=0.7\linewidth]{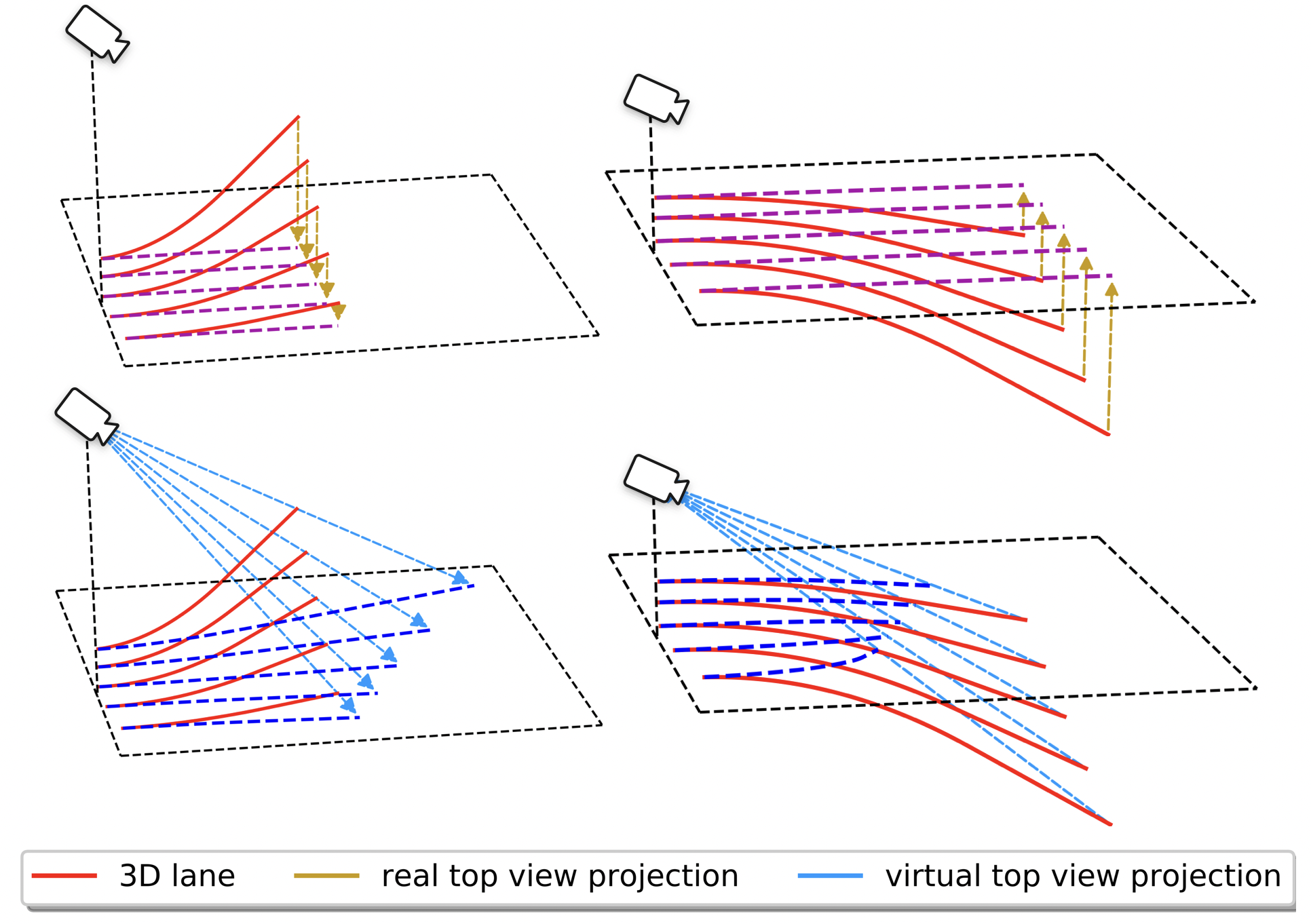}
    \caption{\textbf{Comparison of real and virtual BEV.} Left: Uphill scenario, Right: Downhill scenario. The top row shows the 2D lane representation generated by real top view (BEV) projection~\cite{guo2020gen}, while 2D lanes on the bottom row are generated by virtual top view (BEV) projection~\cite{garnett20193d}. The 2D lane representation generated by virtual BEV projection~\cite{guo2020gen} could reflect the changes of lane height in 3D space.}
    \label{fig:top_projection}
\vspace{-0.4cm}
\end{figure}

\par {\textbf{View projection.} In 3D-LaneNet~\cite{garnett20193d}, a real BEV projection is utilized for creating lane anchors on the flat ground. Real BEV stands for the direct vertical projection from 3D space to the ground plane $g$. In this case, for a point $P_{3D} (x_{3D},y_{3D},z)$ in 3D space, the height dimension $z$ is simply discarded, and the point is projected onto the ground plane position of 
$P_{{2D}R} (x_{3D},y_{3D})$ under real BEV projection with center at vehicle ego coordinate. However, as shown in the top row of Figure~\ref{fig:top_projection}, such 2D lane representation cannot properly reflect the change of lane height in 3D space, thus a heavy image feature encoder in \cite{garnett20193d} is necessary to estimate lane height from images.}

Gen-LaneNet~\cite{guo2020gen} then introduces the virtual BEV projection. As shown in the bottom row of Figure~\ref{fig:top_projection}, 3D lanes are projected onto the ground plane via virtual BEV projection, which can be obtained from rays start from camera center $C$ to the ground $g$. This is conceptually equivalent to 1) project the 3D lanes onto the image plane, and then 2) project image and lanes onto the ground plane by IPM. In this case, the 2D representation of lanes is no longer agnostic of height variance. Equation~\ref{eq:virtual_top} shows the transformation of lane point coordinates from point $P_{3D}$ in 3D to point $P_{{2D}V}$ on the ground plane via virtual BEV projection. With the increment of lane height $z$, the $x$ and $y$ of lane points on ground plane $g$ would be projected away from the positions on real BEV, causing the lane boundaries to be divergent in the uphill scenario.

\begin{equation}
\centering
\begin{split}
P_{{2D}V}=\overrightarrow{C P_{3D}}\cap \Pi_{g}=\frac{h_{cam}}{h_{cam}-z} \cdot
\begin{pmatrix}
x_{3D} \\
y_{3D}
\end{pmatrix}
\label{eq:virtual_top}
\end{split}
\end{equation}

\par {\textbf{Geometry prior.} Parallel lane boundaries and constant lane width are basic assumptions for nearly all lane-based applications such as lane centering assistant (LCA) and lane keeping assistant (LKA). 
Instead of making a strong assumption that multiple lanes in one frame share a global lane width\cite{nieto2008robust, jiang2010computer}, it is common to assume that a single lane would keep a relatively fixed width as it extends to infinity. 
For flat ground cases, the 2D lane representation projected via the virtual BEV would be parallel and have constant lane width. 
In reality, not-flat ground cases are not rare, such as twisted lanes on a helicoidal surface. In this case, parallelism of the projected lane boundaries is not satisfied. Thus, the basic assumption of parallelism and constant width\cite{dementhon1987zero} can only be established in 3D space.}

Moreover, as shown in Equation~\ref{eq:virtual_top}, under virtual BEV projection, lane boundaries in 3D space would be mapped onto the flat ground in different scales w.r.t. height information. Conceptually, the curvature of mapped lane boundaries is positively correlated to the lane slope in 3D space. That is to say, the width change in 2D projection could reflect the variance of lane height in 3D. This provides basic theoretical intuition of how the height information in 3D can be reconstructed from a monocular 2D image, and the geometry structure in the 2D lane mask can be the guidance for estimating the 3D lanes in the real world.

\begin{figure}[t]
    \centering
    \includegraphics[width=0.65\linewidth]{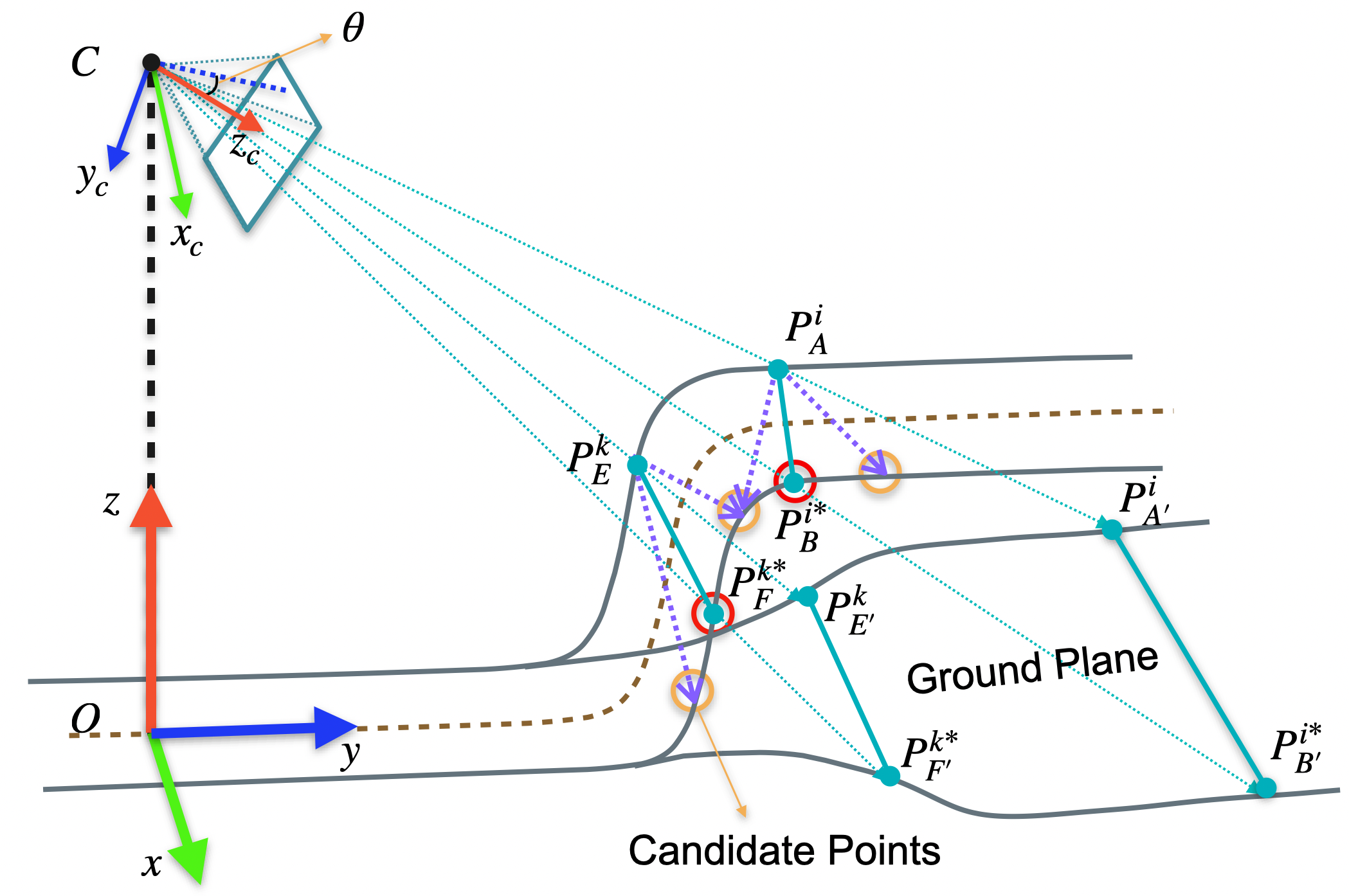}
    \caption{\textbf{Geometry prior of 3D lanes.}}
    \label{fig:lane_width}
\vspace{-0.4cm}
\end{figure}

\par {\textbf{Geometry prior guided supervision.}}
As illustrated above, the encoding of geometry structure information plays a vital role in accurately reconstructing lanes in 3D space. Specifically, we focus on the intra-lane and inter-lane properties between 3D lanes and 2D representation under the virtual BEV projection. We propose the \textit{geometry prior loss}, an auxiliary loss function to involve explicit supervision in the local-to-global preservation of 3D lane structure.

First, we build constraints upon the generalizable geometry prior of lane width in 3D space. The 3D lane width can be calculated as the 3D Euclidean distance $D_{3D}$ between the closest point pair $P_{l}^{i}$ on the left lane boundary and $P_{r}^{i*}$ on the right as $D_{3D}(P_{l}^{i},P_{r}^{i*})=|\overrightharpoonup{P_{l}^{i}P_{r}^{i*}}|$, where $i$ and $i*$ denote the corresponding anchor order in their y-anchor representation.
Besides, we apply a greedy matching algorithm with sliding windows to find the corresponding lane width pairs (LWP) of points from the left lane boundary to the right. Formally: $LWP\{i,i*\}=\text{argmin}_{j}^{}(D_{3D}(P_{l}^{i},P_{r}^{j})); 
j\in\mathbb{N} : \{i-1,i,i+1\}$

As shown in Figure~\ref{fig:lane_width}, distance $D_{3D}(P_{A}^{i},P_{B}^{i*})$ and $D_{3D}(P_{E}^{k},P_{F}^{k*})$ should have the same magnitude. For a specific lane instance, we calculate $D_{3D}$ for all point pairs on the lane boundaries with visibility $\hat{v_{l}}=1$  as in Equation~\ref{eq:3D_distance_loss}. The visibility flag $\hat{v}$ indicates whether a 3D lane point is visible when projected onto the Front view image.
\begin{equation}
\begin{split}
D_{3D}(P_{l}^{i},P_{r}^{i*})&=D_{3D}(P_{l}^{(i+1)},P_{r}^{(i+1)*}); \\
&i\in\mathbb{N} : \{0,1,2,...\sum(\mathbbm{1}\{\hat{v_{l}^{}}=1\})-1\}
\label{eq:3D_distance_loss}
\end{split}
\end{equation}

Second, since projection method makes it achievable to bridge the 2D representation and 3D lanes in an explicit manner, we propose to make a joint optimization of offsets $x$, $y$, and height $z$ by their relationship on the ground plane.
As illustrated in Figure~\ref{fig:lane_width}, point pairs $\{P_A',P_B'\}$ and $\{P_E',P_F'\}$ on ground plane are from the virtual BEV projection of 3D point pairs $\{P_A,P_B\}$ and $\{P_E,P_F\}$. We define the 2D distance $D_{2D}$ as the 2D Euclidean distance weighted by camera height $h_{cam}$ and lane height. For the cases when $z_{l}^{i}$ and $z_{r}^{i*}$ are not the same, we use their 
mean value $\tilde{z}$ in approximation of the nonlinear function $D_{2D}$.

\begin{equation}
\begin{split}
D_{2D}(P_{l}^{i},P_{r}^{i*})=&\sqrt{(x_{l}^{i}-x_{r}^{i*})^2+(y_{l}^{i}-y_{r}^{i*})^2}\cdot (h_{cam}-\tilde{z});\\
&\tilde{z} \sim (z_{l}^{i}+z_{r}^{i*})/2
\label{eq:2D_distance_1}
\end{split}
\end{equation}
If and only if the point pairs $\{P_A',P_B'\}$ and $\{P_E',P_F'\}$ share the same height respectively, we have 

\begin{equation}
\begin{split}
&D_{2D}(P_{l}^{i},P_{r}^{i*})=D_{3D}(P_{l}^{i},P_{r}^{i*})*h_{cam}= \\
&D_{2D}(P_{l}^{(i+1)},P_{r}^{(i+1)*})=D_{3D}(P_{l}^{(i+1)},P_{r}^{(i+1)*})*h_{cam}
\label{eq:2D_distance_2}
\end{split}
\end{equation}
in which the 2D distance equals to the product of 3D lane Euclidean distance and the camera height for $i\in\mathbb{N}: [0,\sum(\mathbbm{1}\{\hat{v_{l}^{}}=1\})-1]  \iff(z_{l}^{i}=z_{r}^{i*}) \cap (z_{l}^{(i+1)}=z_{r}^{(i+1)*})$

For twisted lanes, we relax the constraint in Equation~\ref{eq:2D_distance_2}. Under the heuristic that the curves of lane boundaries in 3D space have continuous derivatives in most cases, the height difference $\Delta (z_{l}- z_{r})$ between point pairs is also locally smooth. 
As the $D_{2D}$ is respect to the height difference of lane point pairs, specifically, we have Equation~\ref{eq:2d_distance_loss} for the point pair ${i,i*}$  and its neighbor pairs in the y-direction on the same lane.
\begin{equation}
\begin{split}
&D_{2D}(P_{l}^{(i-1)},P_{r}^{(i-1)*})\leq D_{2D}(P_{l}^{i},P_{r}^{i*})\leq \\
&D_{2D}(P_{l}^{(i+1)},P_{r}^{(i+1)*}); 
i \in\mathbb{N}:[0,N-1] \Rightarrow \exists \frac{\partial D_{2D}}{\partial y}
\label{eq:2d_distance_loss}
\end{split}
\end{equation}

Finally, we impose the dynamic constraint to the point pairs of each predicted lane with supervision from their neighboring pairs, as shown in Equation~\ref{eq:geo_loss}. For $d \in \{2D,3D\}$, such supervision is applied on the predicted lane with probability $p$ and visibility $v$ equal to 1 in the corresponding ground truth (GT).
\begin{equation}
\begin{split}
L_{geo}= \sum_{d}^{}\sum_{i=1}^{N-1}p\cdot(||v_{}^{i}\cdot((D_{d}^{i-1}+D_{d}^{i+1})-2*D_{d}^{i})||_1)
\label{eq:geo_loss}
\end{split}
\end{equation}
The definition of $L_{geo}$ provides auxiliary supervision incorporating geometry prior into the training pipeline, which also leaves a margin for natural noises and gradual changes in lane width by weakening the geometry assumption of fixed lane width to adapt to both synthetic and real cases. The optimization of $L_{geo}$ guides the model to maintain the lane structure from local to global.

\begin{figure}
    \centering
    \includegraphics[width=\linewidth]{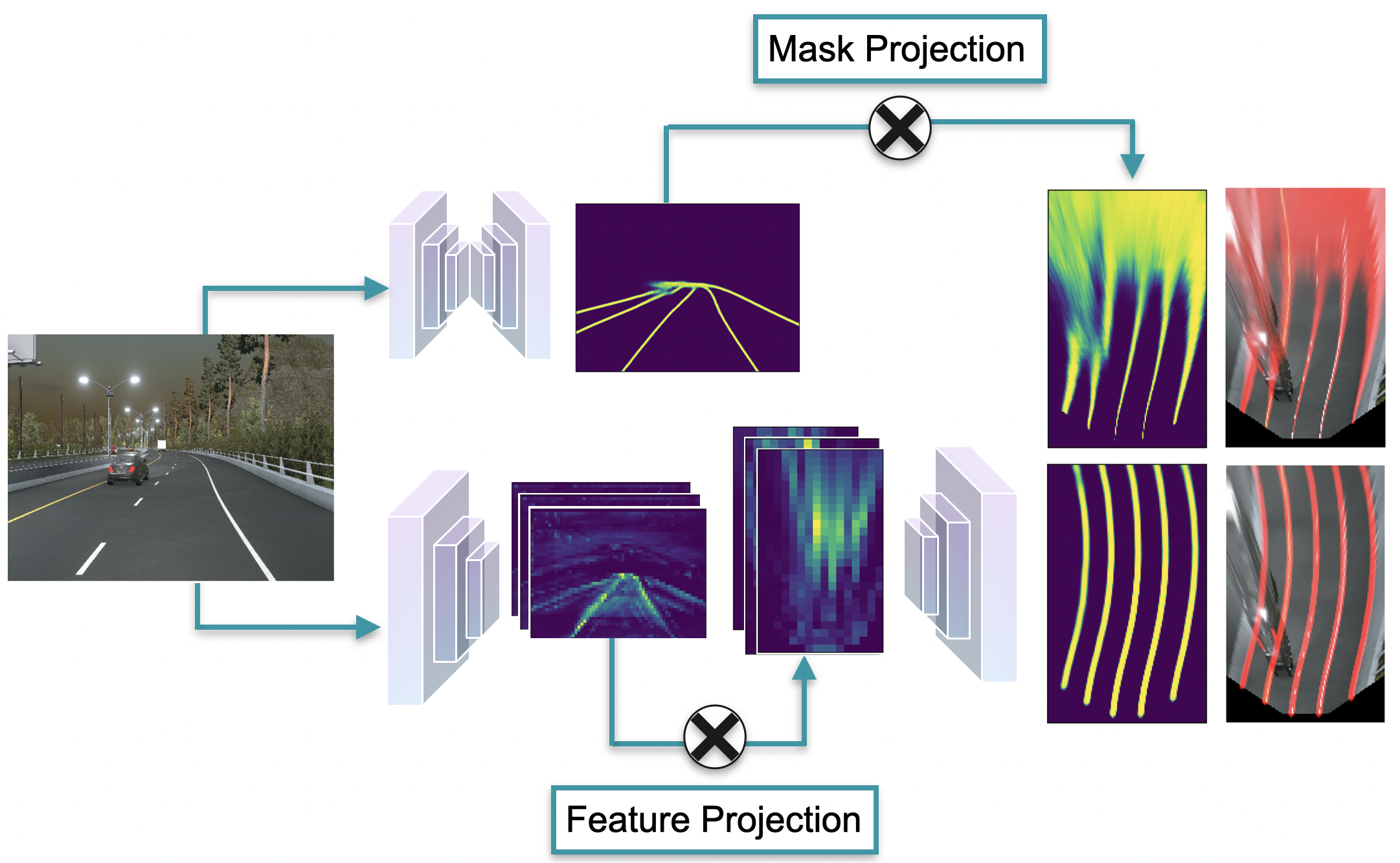}
\caption{\textbf{Comparison between front view supervision with mask projection~\cite{guo2020gen}
and our BEV supervision with feature projection.}}
    \label{fig:top_view_super}
\vspace{-0.6cm}
\end{figure}

\subsection{Supervision from BEV}
As illustrated in section \hyperref[sec:geoprior]{3.1}, the structure of 3D can be discerned from the geometry structure underlying the 2D lane mask. However, structure information like lane width variance is very subtle thus requiring accurate extraction of 2D lanes feature from the front view image.

In Gen-LaneNet\cite{guo2020gen}, front view GT masks are used as supervision for the segmentation network. In the regression network, the mask is first projected onto the ground plane via planar IPM, and then fed into a lane regression network for estimating the 3D lanes. However, as shown in the top row of Figure~\ref{fig:top_view_super}, when the mask is projected onto the ground plane, the features on the far side are blurry, and different lane features are mixed together, which prevent the subsequent network from accurately decoding the structure information contained in the feature maps. We argue that the distant lane feature confusion is due to the misleading mask supervision conducted in the front view for the following reasons.

First, the IPM warps the front view features onto the ground plane via a non-linear transformation. And it is hard for the front view segmentation network to distinguish far lane instances in pixel level since the lanes are converging to the vanishing point and overlapped with each other. As shown on the top right of Figure~\ref{fig:top_view_super}, few far pixels on the front view will be mapped into a huge range on BEV, so the projected segmentation mask will be stretched drastically in the far end, resulting in feature confusion. 
Second, positive lane boundary pixels in the ground truth masks of Gen-LaneNet\cite{guo2020gen} are generated on the front view with equal line thickness. However, due to the perspective effect, lane markings are always thicker in the close end and thinner in the far end. As a result, this kind of ground truth mask would not align properly with the shape of lane markings, which will lead to a fluctuation in extracting 2D lane structure and decoding 3D geometry information.

To address these issues, we propose a refined pipeline in which the ground truth mask is generated and used as supervision on BEV, as shown in the bottom row of Figure~\ref{fig:top_view_super}. We first project lane points from the front view to the virtual BEV via the homography matrix $H$, then generate the GT mask of lane boundaries with equal thickness. 
A projective transformation network \cite{garnett20193d,jaderberg2015spatial} is used to project the output features of image encoder to BEV, and conduct segmentation supervision directly on BEV mask from decoder by optimizing the standard cross-entropy loss $L_{seg}$. Besides, we can also jointly estimate camera pitch $\theta$ and height $h_{cam}$ from the front view images with $L_{cam}=||\theta-\hat{\theta}||_1+|h_{cam}-\hat{h}_{cam}||_1$. Such direct mask supervision in BEV will tremendously resolve the problem of feature confusion and preserve the global lane geometry structure information in BEV 2D mask. 

\begin{table*}
\caption{Comprehensive evaluation on Apollo 3D synthetic dataset\cite{guo2020gen}. Parameters: Estimated network parameters to the best of our knowledge. GT mask: Use GT masks as the input of lane regression network to test the theoretical upper bound of corresponding method. The extra-long range refers to the data split of range 0-200m as described in section \hyperref[sec:extra]{4.3}, while other splits refer to the default evaluation range of 0-100m.} 
    \centering
    \label{tab:main}
    \resizebox{0.72\textwidth}{!}{
    \begin{tabular}{lc|cc|cc|cc|cc}
    \hline
    \multicolumn{2}{c|}{}                     & \multicolumn{2}{c|}{balanced scenes} & \multicolumn{2}{c|}{rarely observed} & \multicolumn{2}{c|}{visual variations}                             & \multicolumn{2}{c}{{\textbf{extra-long range}}}  \\ \hline
    \multicolumn{1}{l|}{Method}                & parameters & F-Score           & AP               & F-Score           & AP               & F-Score           & AP                               & F-Score           & AP                   \\ \hline
    \multicolumn{1}{l|}{3D-LaneNet\cite{garnett20193d}}            & 23.7M        & 86.4              & 89.3             & 72.0              & 74.6             & 72.5              & 74.9         & 60.1              & 63.2                 \\
    \multicolumn{1}{l|}{Gen-LaneNet\cite{guo2020gen}}           & 2.8M       & 88.1              & 90.1             & 78.0              & 79.0             & 85.3              & 87.2            & 68.5              & 69.2                 \\
    \multicolumn{1}{l|}{Ours}                  & 2.8M       & \textbf{91.9}     & \textbf{93.8}    & \textbf{83.7}     & \textbf{85.2}    & \textbf{89.9}     & \textbf{92.1}                    & \textbf{83.6}    & \textbf{85.3}                 \\ \hline
    \multicolumn{1}{l|}{Gen-LaneNet (GT mask)\cite{guo2020gen}} & -          & 91.8              & 93.8             & 84.7              & 86.6             & 90.2              & 92.3            & 80.7              & 82.5                 \\
    \multicolumn{1}{l|}{Ours (GT mask)}        & -          & \textbf{92.8}     & \textbf{94.7}    & \textbf{87.8}     & \textbf{89.5}    & \textbf{91.3}     & \textbf{93.2}                    & \textbf{87.2}    & \textbf{89.1}                 \\ \hline
    \end{tabular}}
\end{table*}

\begin{table*}
    \caption{Results on offset metric and ablation study on balanced scenes in Apollo 3D synthetic dataset\cite{guo2020gen}. GS: Geometry supervision. BVS: BEV supervision. Aug: 3D lane data augmentation. GT mask: Use ground truth mask of 2D lanes instead of predicted mask for 3D lane regression. Joint: Evaluate on the intersection of matched samples across all methods in the table.} 
    \centering
    \label{tab:offset}
    \resizebox{0.95\textwidth}{!}{
    \begin{tabular}{l|cccccccc}
    \hline
    Method                                  & F-Score            & AP                  & x error near (m)       & x error far (m)         & {\textbf{joint}} x error far (m) & z error near (m)     & z error far (m)    & {\textbf{joint}} z error far (m) \\ \hline
    3D-LaneNet \cite{garnett20193d}         & 86.4               & 89.3                & 0.068                  & 0.477                   & 0.448                 & 0.015                & 0.202                           & 0.183                    \\
    Gen-LaneNet     \cite{guo2020gen}\textbf{(baseline)}       & 88.1               & 90.1                & 0.061                  & 0.496                   & 0.470                 & 0.012                & 0.214                           & 0.192                    \\
    Ours w/ GS                              & 89.1               & 91.2                & 0.050                  & 0.455                   & 0.406                 & 0.009                & 0.226                           & 0.183                    \\
    Ours w/ BVS                             & 90.2               & 92.4                & 0.060                  & 0.446                   & 0.383                 & 0.013                & 0.241                           & 0.189                    \\
    Ours w/ GS + BVS                        & 91.2               & 93.2                & 0.065                  & 0.415                   & 0.365                 & 0.009                & 0.220                           & 0.179                    \\
    Ours w/ GS + BVS + Aug                  & \textbf{91.9}      & \textbf{93.8}       & 0.049                  & 0.387                   & \textbf{0.332}        & 0.008                & 0.213                           & \textbf{0.171}                    \\ \hline
    Gen-LaneNet (GT mask) \cite{guo2020gen} & 91.8               & 93.8                & 0.054                  & 0.412                   & 0.353                 & 0.011                & 0.226                           & 0.177                    \\
    Ours (GT mask)                          & \textbf{92.8}      & \textbf{94.7}       & 0.044                  & 0.360                   & \textbf{0.299}        & 0.007                & 0.219                           & \textbf{0.170}                    \\ \hline
    \end{tabular}}
    
\end{table*}

\begin{table}
    \centering
    \caption{Comparison of augmentation on easy and hard samples}
    \label{tab:aug}
    \resizebox{1.0\linewidth}{!}{
    \begin{tabular}{l|cc|cc|cc}
    \hline
    \multicolumn{1}{c|}{} & \multicolumn{2}{c|}{all}        & \multicolumn{2}{c|}{easy}      & \multicolumn{2}{c}{hard}            \\ \hline
    Method                & F-Score        & AP             & F-Score       & AP             & F-Score           & AP              \\ \hline
    Ours w/ GS              & 89.1                      &91.2                           & 93.2              & 92.7              & 69.3                  & 77.7 \\
    Ours w/ BVS             & 90.2                      & 92.4                          & 93.4              & 94.2              & 70.5                  & 78.7  \\
    Ours (GS+BVS w/o Aug)        & 91.2           &93.2            & 94.2          & 95.0           & 71.9              & 79.1            \\
    Ours (GS+BVS w/ Aug)         & \textbf{91.9} {\color{blue}(+0.7)} & \textbf{93.8} {\color{blue}(+0.6)}  & \textbf{94.6} {\color{blue}(+0.4)} & \textbf{95.2} {\color{blue}(+0.2)}  & \textbf{75.4} {\color{blue}(+3.5)} & \textbf{83.7} {\color{blue}(+4.6)}   \\ \hline
    \end{tabular}}
\end{table}

\subsection{Lane Augmentation in 3D Space}

In both synthetic dataset and real world, the vast majority of ground planes are flat with almost zero slope, which severely dominates the distribution of training data. According to our statistics, in the Apollo synthetic dataset~\cite{guo2020gen}, 67.8\% of lanes in training data have lane height within the range of [-0.1,0.1]m, while 44.4\% of lanes fall into the range of [-0.01,0.01]m. Even though such a biased distribution is a close reflection of reality, it would dominate the direction of gradient optimization away from the long tail data, such as uphill and downhill cases.

\begin{figure}
    \centering
    \includegraphics[width=0.9\linewidth]{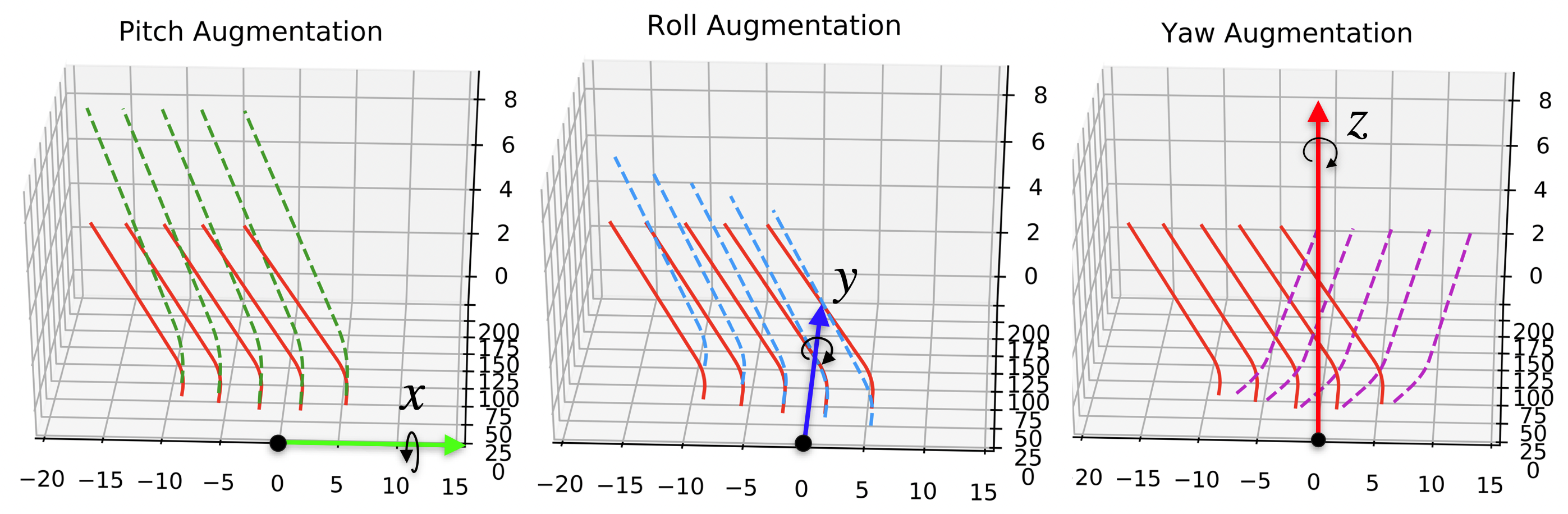}
    \caption{\textbf{Pitch, Roll and Yaw rotation augmentation on 3D lanes.}}
    \label{fig:augmentation}
\vspace{-0.4cm}
\end{figure}

Conventional augmentation methods, such as translation or center rotation, would violate the geometry prior that 3D lanes start from the center of ground plane. Thus, we propose a novel 3D lane rotation augmentation method to deal with the  long tail distribution. Figure~\ref{fig:augmentation} shows the process of generating augmentation data by applying pitch, yaw and roll rotation on the 3D lanes. An example of applying pitch augmentation on 3D point $P$ to get the target point $P'$ under homogeneous transformation is shown in Equation~\ref{eq:aug}. The rotated 3D points are projected onto front view to generate augmented input signals, which is equivalent to applying a perspective augmentation on the image and would be beneficial to the jointly understanding of semantic mask and camera pose.

\begin{equation}
P'=\textbf{R}_x P ;
\textbf{R}_x = 
\begin{bmatrix}
1 & 0 & 0 \\
0 & cos \theta & -sin \theta \\
0 & sin \theta & cos \theta
\end{bmatrix}
\label{eq:aug}
\end{equation}

By applying pitch rotation, 
the ground plane would have greater fluctuation of slope while preserving the inner structure of lanes.
By applying yaw rotation, 
the lane patterns are enriched without affecting the lane height. 
Lastly, by applying roll rotation, twisted 3D lanes are generated. 
Experimental results in Table~\ref{tab:aug} show the effectiveness of 3D lane augmentation over hard samples, while the results in section \hyperref[sec:extra]{4.3} shows its superiority on extra-long range data. Our proposed method can also be applied in both stages of mask prediction and anchor-based regression to improve robustness by building feature consistency under different perspective transformation~\cite{zhao2021camera, zhou2021monocular}.

\subsection{Training}
Given an RGB image and the corresponding 3D lanes, a detailed pipeline for the framework is shown in Figure~\ref{fig:flowchart}.
For the segmentation head in feature extraction network, we optimize the cross-entropy loss $L_{seg}$. And for the optional camera pose regression head connected to the front-view image encoder, we apply the $L_{1}$-loss for supervision as in $L_{cam}$. The input of lane reconstruction network is the logits from the feature extraction network. Following the lane anchor settings in 3D-LaneNet\cite{garnett20193d} and Gen-LaneNet\cite{guo2020gen}, lanes in 3D space are first projected onto the flat ground with virtual BEV projection, then associated with closest anchor at $Y_{ref}$. For a predicted anchor $X_{A}^{i}$ with probability $p$ and the ground truth anchor $\hat{X_{A}^{i}} = \{(\hat{x_{i}^{}},\hat{z_{i}^{}},\hat{v_{i}^{}},\hat{p})\}$, each anchor point at a pre-defined y position ${y}_{i}$ predicts the x offset $x_{i}$, the lane height $z_{i}$ and the visibility $v_{i}$ which means whether the 3D points are visible when projected onto front view. Both x-offset and y-position are defined on flat ground plane. We have the anchor loss written as 

\begin{equation}
\vspace{-0.4cm}
\begin{split}
L_{anchor}&=-\sum_{i=1}^{N}(\hat{p}_{i}\log{p_{i}}+(1-\hat{p}_{i})(\log{(1-p_{i}}))\\
&+\sum_{i=1}^{N}\hat{p}_{i}\cdot(||\hat{v}_{i}\cdot(x_{i}-\hat{x}_{i})||_1+||\hat{v}_{i}\cdot(z_{i}-\hat{z}_{i})||_1)\\
&+\sum_{i=1}^{N}\hat{p}_{i}\cdot ||v_{i}-\hat{v}_{i} ||_{1}
\label{eq:anchor_loss}
\end{split}
\end{equation}

To sum up, the loss functions are composed of the loss of BEV feature extraction network $L_{fea}$ and the loss of 3D lane reconstruction network $L_{rec}$.
\begin{equation}
\begin{split}
&L_{fea}=L_{seg}+\lambda_{cam}L_{cam}\\
&L_{rec}=L_{anchor}+\lambda_{geo}L_{geo}
\label{eq:all_loss}
\end{split}
\end{equation}
where the $\lambda_{cam}$ and $\lambda_{geo}$ are introduced to balance the relative importance between different loss items.

\begin{figure*}
    \centering
    \includegraphics[width=0.65\linewidth]{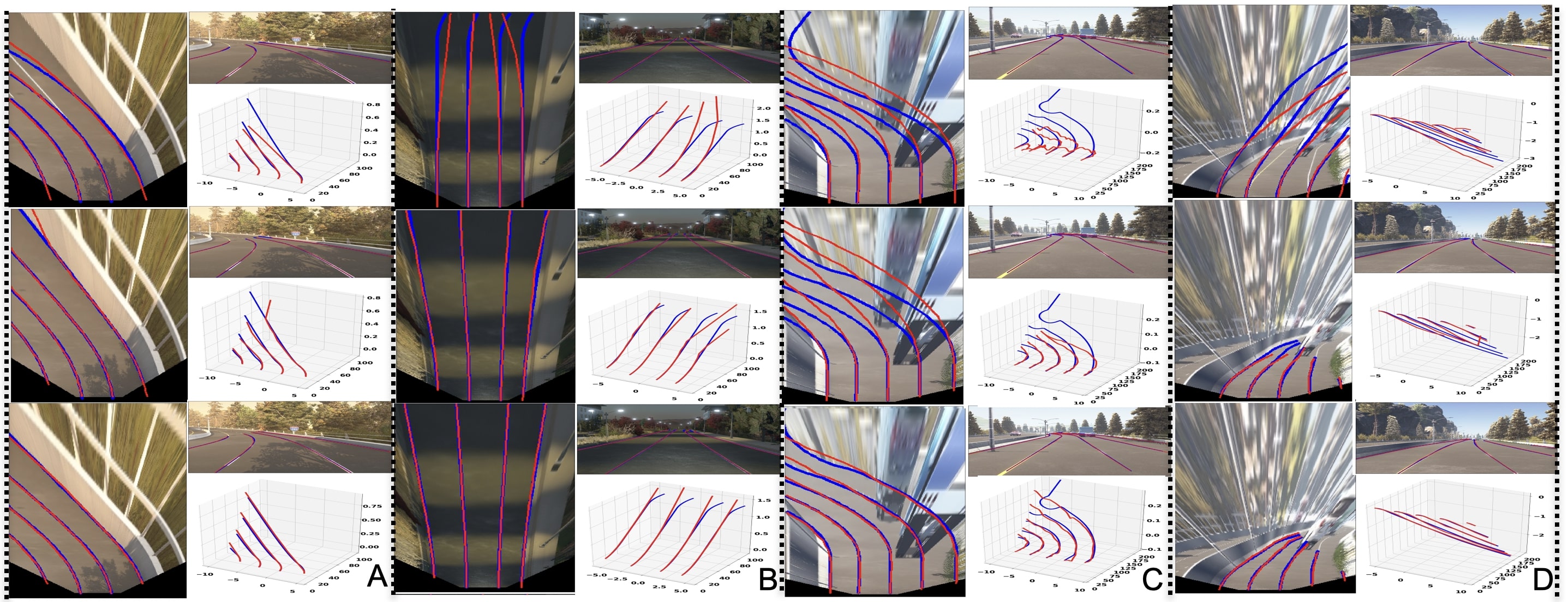}
    \caption{\textbf{Qualitative comparison results of proposed method.} First row: 3D-LaneNet~\cite{garnett20193d}; Second row: Gen-LaneNet~\cite{guo2020gen}; Third row: Our proposed method;different GTs are generated from different view projection; Column A \& B: Results on default range (0-100m); Column C \& D: Results on extra-long range (0-200m). GT and predicted lane boundaries are marked in blue and red respectively. Our proposed method could tremendously facilitate the reconstruction of 3D lanes, especially on the far side.}
    \label{fig:vis}
\vspace{-0.4cm}
\end{figure*}

\section{Experiment}
\label{sec:Experiment}
\subsection{Datasets and Implementation Details}
To validate the performance of our proposed framework, we adopt the Apollo 3D synthetic dataset~\cite{guo2020gen} for comparative experiments since other 3D lane datasets are not publicly available. 
This dataset contains 10500 discrete frames of monocular RGB images and the corresponding lane keypoint labels in 3D world coordinate. Follow the same setting in previous literature~\cite{guo2020gen}, the data is split into balanced, rarely observed and visual variation scenes.
For evaluation metric, we adopt the minimum-cost flow algorithm \cite{guo2020gen} for matching between GT and predicted lanes, and report Average Precision (AP) and F-score for comprehensive evaluations. Besides, we also report horizontal offset on flat ground and the height distance in meters between matched pairs of GT and prediction lane on near (0-40m) and far (40-100m) range. We have $\lambda_{cam}$=1$e^{2}$ and $\lambda_{geo}$=1$e^{-2}$ respectively.

For anchor representation, we adopt y-reference points in \{5, 10, 15, 20, 30, 40, 50, 60, 80, 100\}m, and set up $Y_{ref}$ = 5 to associate each lane label with the closest anchor. For model training, we adopt the Adam~\cite{kingma2014adam} optimizer  with initial learning rate of 5$e^{-4}$, and no scheduler or weight decay is applied throughout the training. Our network can be trained in an end-to-end or two-stage way. We use batch size of 8 and train the entire network for 50 epochs on one GeForce GTX 1080Ti GPU. We adopt the same network architecture as ERFNet~\cite{romera2017erfnet} and GeoNet in~\cite{guo2020gen}. For lane augmentation, we randomly generate new lane data applying rotation for pitch  within [-0.1,0.3] for $p_{pitch}$ = 0.1, roll within [-3,3] for $p_{roll}$ = 0.05, and yaw within [-3,3] for $p_{yaw}$ = 0.2. We assume known camera pose in the training for fair comparison with previous work, even though the pose information can be obtained by network prediction as well.

\subsection{Evaluation Result}
The comprehensive experiment results are shown in Table~\ref{tab:main}, and the qualitative comparison is shown in Figure~\ref{fig:vis}. Compared with the previous state-of-the-art works, we increase the F-score for more than 3.8\% on all data splits without introducing extra network parameters, and our prediction results can closely achieve the theoretical results of Gen-LaneNet~\cite{guo2020gen} with GT mask for the segmentation network. We also testify our methods on regression network with GT masks as input, and show that our framework can tremendously reduce the discrepancy between results under predicted 2D lane masks and GT masks with the same model architecture as~\cite{guo2020gen}, which proves the promising effect of our method in refining BEV features and preserving 3D lane structure.

\subsection{Extra-long Range Comparison}
\label{sec:extra}
The ability of detecting lanes at long distance is critical especially in the high speed scenario, thus we design an ``extra-long range'' data split to evaluate the ability of detecting 3D lanes within the range 0-200m, doubling the original evaluation range. We create this split by filtering lane data from the balanced scenes in~\cite{guo2020gen} with y-position greater than 195m, and equally sample y-reference points from range [5, 200]m with a 5-meter interval. Experiment results are shown in Table~\ref{tab:main}.  With the refinement of far-side features and the explicit geometry supervision, our proposed methods can increase both AP and F-score for roughly 15\% over~\cite{guo2020gen}  and 25\% over ~\cite{garnett20193d}. Specifically, our augmentation method makes an improvement on F-score from 79.7\% to 83.6\% and AP from 81.2\% to 85.3\%, which proves the superiority for generalizing into long-range detection.

\subsection{Ablation Study}
We conduct ablation study for evaluating the impact of each major part of our proposed method: Geometry supervision (GS), BEV supervision (BVS) and 3D lane augmentation (Aug). We notice that in the evaluation of the offset metrics, the number of samples (i.e., positive matched lanes) evaluated in each method is not exactly the same, thus fairness cannot be assured since methods with higher F-score will consider more matched pairs of lanes which are often hard samples. To ensure fair comparison, we design a ``joint'' offset metric in which the offset errors are evaluated with the intersection of matched data samples from all methods in the table. 
As shown in Table~\ref{tab:offset}, we use Gen-LaneNet~\cite{guo2020gen} as the baseline. The GS and BVS can increase the performance of the baseline respectively. Compared with only applying GS on the baseline, the joint offset errors in the row 3-6 show that applying GS on the refined features with BVS can result in a much greater improvement in refining both ``x error far'' and ``z error far'', which reflects the joint benefit of GS and BVS on the preservation of geometry information in reconstructing lanes from 2D to 3D. Besides, to testify the ability to generalize on rare cases for our lane augmentation method, we split the balanced scenes of \cite{guo2020gen} into easy and hard cases. An image is labeled as hard case if the lane height of any lane within the image is greater than 1.78m, which is the value of the shared camera height in the dataset \cite{guo2020gen}. In general, 184 out of 1496 images are labeled as hard cases and the other 1312 images are labeled as easy. As shown in Table~\ref{tab:aug}, 
the proposed augmentation method could tremendously improve the model performance on hard cases with 4.6\% increment on AP and 3.5\% on F-score, which proves the effectiveness on countering imbalanced data distribution.

\section{Conclusion}
In this paper\footnote{For the discussion of limitations and future work, please refer to the supplementary material.}, we propose a framework for monocular 3D lane detection with explicit geometry supervision, which outperforms previous methods without introducing extra parameters. We consider the task as a 2D to 3D lane reconstruction problem and prove that the structure prior could tremendously facilitate the reconstruction of 3D lanes by offering explicit guidance for lane geometry and refined lane features. On the Apollo 3D synthetic dataset, our proposed method can improve the performance of state-of-the-art methods by a large margin. Besides, our proposed approaches show the potential for extra-long distance 3D lane detection and achieve roughly 15\% improvement in F-Score and AP over existing methods on the long-range data split.

{\small
\bibliographystyle{ieee_fullname}
\bibliography{egbib}

\begin{thebibliography}{10}\itemsep=-1pt

\bibitem{aly2008real}
Mohamed Aly.
\newblock Real time detection of lane markers in urban streets.
\newblock In {\em 2008 IEEE Intelligent Vehicles Symposium}, pages 7--12. IEEE,
  2008.

\bibitem{bai2018deep}
Min Bai, Gellert Mattyus, Namdar Homayounfar, Shenlong Wang, Shrinidhi~Kowshika
  Lakshmikanth, and Raquel Urtasun.
\newblock Deep multi-sensor lane detection.
\newblock In {\em 2018 IEEE/RSJ International Conference on Intelligent Robots
  and Systems (IROS)}, pages 3102--3109. IEEE, 2018.

\bibitem{bertozzi1998gold}
Massimo Bertozzi and Alberto Broggi.
\newblock Gold: A parallel real-time stereo vision system for generic obstacle
  and lane detection.
\newblock {\em IEEE transactions on image processing}, 7(1):62--81, 1998.

\bibitem{butakov2014personalized}
Vadim~A Butakov and Petros Ioannou.
\newblock Personalized driver/vehicle lane change models for adas.
\newblock {\em IEEE Transactions on Vehicular Technology}, 64(10):4422--4431,
  2014.

\bibitem{chen2017end}
Zhilu Chen and Xinming Huang.
\newblock End-to-end learning for lane keeping of self-driving cars.
\newblock In {\em 2017 IEEE Intelligent Vehicles Symposium (IV)}, pages
  1856--1860. IEEE, 2017.

\bibitem{coulombeau2002vehicle}
Pierre Coulombeau and Claude Laurgeau.
\newblock Vehicle yaw, pitch, roll and 3d lane shape recovery by vision.
\newblock In {\em Intelligent Vehicle Symposium, 2002. IEEE}, volume~2, pages
  619--625. IEEE, 2002.

\bibitem{dementhon1987zero}
Daniel DeMenthon.
\newblock A zero-bank algorithm for inverse perspective of a road from a single
  image.
\newblock In {\em Proceedings. 1987 IEEE International Conference on Robotics
  and Automation}, volume~4, pages 1444--1449. IEEE, 1987.

\bibitem{dickmanns1992recursive}
Ernst~D. Dickmanns and Birger~D. Mysliwetz.
\newblock Recursive 3-d road and relative ego-state recognition.
\newblock {\em IEEE Transactions on Pattern Analysis \& Machine Intelligence},
  14(02):199--213, 1992.

\bibitem{duan2019centernet}
Kaiwen Duan, Song Bai, Lingxi Xie, Honggang Qi, Qingming Huang, and Qi Tian.
\newblock Centernet: Keypoint triplets for object detection.
\newblock In {\em Proceedings of the IEEE/CVF International Conference on
  Computer Vision}, pages 6569--6578, 2019.

\bibitem{efrat20203d}
Netalee Efrat, Max Bluvstein, Shaul Oron, Dan Levi, Noa Garnett, and Bat~El
  Shlomo.
\newblock 3d-lanenet+: Anchor free lane detection using a semi-local
  representation.
\newblock {\em arXiv preprint arXiv:2011.01535}, 2020.

\bibitem{fu2019dual}
Jun Fu, Jing Liu, Haijie Tian, Yong Li, Yongjun Bao, Zhiwei Fang, and Hanqing
  Lu.
\newblock Dual attention network for scene segmentation.
\newblock In {\em Proceedings of the IEEE/CVF Conference on Computer Vision and
  Pattern Recognition}, pages 3146--3154, 2019.

\bibitem{garnett20193d}
Noa Garnett, Rafi Cohen, Tomer Pe'er, Roee Lahav, and Dan Levi.
\newblock 3d-lanenet: end-to-end 3d multiple lane detection.
\newblock In {\em Proceedings of the IEEE/CVF International Conference on
  Computer Vision}, pages 2921--2930, 2019.

\bibitem{guo2020gen}
Yuliang Guo, Guang Chen, Peitao Zhao, Weide Zhang, Jinghao Miao, Jingao Wang,
  and Tae~Eun Choe.
\newblock Gen-lanenet: A generalized and scalable approach for 3d lane
  detection.
\newblock In {\em Computer Vision--ECCV 2020: 16th European Conference,
  Glasgow, UK, August 23--28, 2020, Proceedings, Part XXI 16}, pages 666--681.
  Springer, 2020.

\bibitem{he2016accurate}
Bei He, Rui Ai, Yang Yan, and Xianpeng Lang.
\newblock Accurate and robust lane detection based on dual-view convolutional
  neutral network.
\newblock In {\em 2016 IEEE Intelligent Vehicles Symposium (IV)}, pages
  1041--1046. IEEE, 2016.

\bibitem{homayounfar2019dagmapper}
Namdar Homayounfar, Wei-Chiu Ma, Justin Liang, Xinyu Wu, Jack Fan, and Raquel
  Urtasun.
\newblock Dagmapper: Learning to map by discovering lane topology.
\newblock In {\em Proceedings of the IEEE/CVF International Conference on
  Computer Vision}, pages 2911--2920, 2019.

\bibitem{jaderberg2015spatial}
Max Jaderberg, Karen Simonyan, Andrew Zisserman, et~al.
\newblock Spatial transformer networks.
\newblock {\em Advances in neural information processing systems},
  28:2017--2025, 2015.

\bibitem{jiang2010computer}
Yan Jiang, Feng Gao, and Guoyan Xu.
\newblock Computer vision-based multiple-lane detection on straight road and in
  a curve.
\newblock In {\em 2010 International Conference on Image Analysis and Signal
  Processing}, pages 114--117. IEEE, 2010.

\bibitem{jin2021robust}
Yujie Jin, Xiangxuan Ren, Fengxiang Chen, and Weidong Zhang.
\newblock Robust monocular 3d lane detection with dual attention.
\newblock In {\em 2021 IEEE International Conference on Image Processing
  (ICIP)}, pages 3348--3352. IEEE, 2021.

\bibitem{kim2008robust}
ZuWhan Kim.
\newblock Robust lane detection and tracking in challenging scenarios.
\newblock {\em IEEE Transactions on intelligent transportation systems},
  9(1):16--26, 2008.

\bibitem{kingma2014adam}
Diederik~P Kingma and Jimmy Ba.
\newblock Adam: A method for stochastic optimization.
\newblock {\em arXiv preprint arXiv:1412.6980}, 2014.

\bibitem{liu2016ssd}
Wei Liu, Dragomir Anguelov, Dumitru Erhan, Christian Szegedy, Scott Reed,
  Cheng-Yang Fu, and Alexander~C Berg.
\newblock Ssd: Single shot multibox detector.
\newblock In {\em European conference on computer vision}, pages 21--37.
  Springer, 2016.

\bibitem{mallot1991inverse}
Hanspeter~A Mallot, Heinrich~H B{\"u}lthoff, JJ Little, and Stefan Bohrer.
\newblock Inverse perspective mapping simplifies optical flow computation and
  obstacle detection.
\newblock {\em Biological cybernetics}, 64(3):177--185, 1991.

\bibitem{nedevschi20043d}
Sergiu Nedevschi, Rolf Schmidt, Thorsten Graf, Radu Danescu, Dan Frentiu,
  Tiberiu Marita, Florin Oniga, and Ciprian Pocol.
\newblock 3d lane detection system based on stereovision.
\newblock In {\em Proceedings. The 7th International IEEE Conference on
  Intelligent Transportation Systems (IEEE Cat. No. 04TH8749)}, pages 161--166.
  IEEE, 2004.

\bibitem{neven2018towards}
Davy Neven, Bert De~Brabandere, Stamatios Georgoulis, Marc Proesmans, and Luc
  Van~Gool.
\newblock Towards end-to-end lane detection: an instance segmentation approach.
\newblock In {\em 2018 IEEE intelligent vehicles symposium (IV)}, pages
  286--291. IEEE, 2018.

\bibitem{nieto2008robust}
Marcos Nieto, Luis Salgado, Fernando Jaureguizar, and Jon Arr{\'o}spide.
\newblock Robust multiple lane road modeling based on perspective analysis.
\newblock In {\em 2008 15th IEEE International Conference on Image Processing},
  pages 2396--2399. IEEE, 2008.

\bibitem{pan2018spatial}
Xingang Pan, Jianping Shi, Ping Luo, Xiaogang Wang, and Xiaoou Tang.
\newblock Spatial as deep: Spatial cnn for traffic scene understanding.
\newblock In {\em Thirty-Second AAAI Conference on Artificial Intelligence},
  2018.

\bibitem{pomerleau1995ralph}
Dean Pomerleau.
\newblock Ralph: Rapidly adapting lateral position handler.
\newblock In {\em Proceedings of the Intelligent Vehicles' 95. Symposium},
  pages 506--511. IEEE, 1995.

\bibitem{redmon2016you}
Joseph Redmon, Santosh Divvala, Ross Girshick, and Ali Farhadi.
\newblock You only look once: Unified, real-time object detection.
\newblock In {\em Proceedings of the IEEE conference on computer vision and
  pattern recognition}, pages 779--788, 2016.

\bibitem{romera2017erfnet}
Eduardo Romera, Jos{\'e}~M Alvarez, Luis~M Bergasa, and Roberto Arroyo.
\newblock Erfnet: Efficient residual factorized convnet for real-time semantic
  segmentation.
\newblock {\em IEEE Transactions on Intelligent Transportation Systems},
  19(1):263--272, 2017.

\bibitem{simonyan2014very}
Karen Simonyan and Andrew Zisserman.
\newblock Very deep convolutional networks for large-scale image recognition.
\newblock {\em arXiv preprint arXiv:1409.1556}, 2014.

\bibitem{tian2019fcos}
Zhi Tian, Chunhua Shen, Hao Chen, and Tong He.
\newblock Fcos: Fully convolutional one-stage object detection.
\newblock In {\em Proceedings of the IEEE/CVF international conference on
  computer vision}, pages 9627--9636, 2019.

\bibitem{xiong20183d}
Lu Xiong, Zhenwen Deng, Peizhi Zhang, and Zhiqiang Fu.
\newblock A 3d estimation of structural road surface based on lane-line
  information.
\newblock {\em IFAC-PapersOnLine}, 51(31):778--783, 2018.

\bibitem{yu2020bdd100k}
Fisher Yu, Haofeng Chen, Xin Wang, Wenqi Xian, Yingying Chen, Fangchen Liu,
  Vashisht Madhavan, and Trevor Darrell.
\newblock Bdd100k: A diverse driving dataset for heterogeneous multitask
  learning.
\newblock In {\em Proceedings of the IEEE/CVF conference on computer vision and
  pattern recognition}, pages 2636--2645, 2020.

\bibitem{zhao2021camera}
Yunhan Zhao, Shu Kong, and Charless Fowlkes.
\newblock Camera pose matters: Improving depth prediction by mitigating pose
  distribution bias.
\newblock In {\em Proceedings of the IEEE/CVF Conference on Computer Vision and
  Pattern Recognition}, pages 15759--15768, 2021.

\bibitem{zhou2021monocular}
Yunsong Zhou, Yuan He, Hongzi Zhu, Cheng Wang, Hongyang Li, and Qinhong Jiang.
\newblock Monocular 3d object detection: An extrinsic parameter free approach.
\newblock In {\em Proceedings of the IEEE/CVF Conference on Computer Vision and
  Pattern Recognition}, pages 7556--7566, 2021.

\end{thebibliography}


\begin{thebibliography}{1}\itemsep=-1pt

\bibitem{garnett20193d}
Noa Garnett, Rafi Cohen, Tomer Pe'er, Roee Lahav, and Dan Levi.
\newblock 3d-lanenet: end-to-end 3d multiple lane detection.
\newblock In {\em Proceedings of the IEEE/CVF International Conference on
  Computer Vision}, pages 2921--2930, 2019.

\bibitem{guo2020gen}
Yuliang Guo, Guang Chen, Peitao Zhao, Weide Zhang, Jinghao Miao, Jingao Wang,
  and Tae~Eun Choe.
\newblock Gen-lanenet: A generalized and scalable approach for 3d lane
  detection.
\newblock In {\em Computer Vision--ECCV 2020: 16th European Conference,
  Glasgow, UK, August 23--28, 2020, Proceedings, Part XXI 16}, pages 666--681.
  Springer, 2020.

\end{thebibliography}
}

\end{document}


\title{Supplementary Material for \\ Reconstruct from BEV: A 3D Lane Detection \\ Approach based on Geometry Structure Prior}

\author{
   {Chenguang Li}\thanks{Both authors contributed equally.}${~~^{1}}$,
   {Jia Shi}\footnotemark[1]
   \thanks{This work was done during internship at SenseTime Research.}${~~^{1,2}}$,
   {Ya Wang}\footnotemark[2]${~~^{1,3}}$,
   {Guangliang Cheng}\thanks{Guangliang Cheng is the corresponding author.}${^{~~1,4}}$ \\
   ${^1}${SenseTime Research} \quad
   ${^2}${Robotics Institute, Carnegie Mellon University} \\
   ${^3}${University of Tuebingen} \quad
   ${^4}${Shanghai AI Laboratory} \\
   {\tt\small \{lichenguang, wangya\}@senseauto.com \, jiashi@andrew.cmu.edu \, guangliangcheng2014@gmail.com}
}

\maketitle

\appendix

\section{Point Pair Searching Algorithm}
\label{sec:supalgo}
In this section, we present the searching algorithm on finding lane point pairs to impose geometry supervision, which is implemented as a greedy matching algorithm with linear time complexity. As shown in Algorithm~\ref{alg:match}, the input of this method is two lists of points on the corresponding neighbour lane boundaries, and the output is the matched key-value pairs of  lane keypoints from the shorter lane boundary to the longer one.

\begin{algorithm}[t] 
\begin{algorithmic}[1]\small
\caption{Search point pairs between two lane boundaries using greedy matching with a sliding window}
\label{alg:match}
\REQUIRE{lane boundary $L1$ with 3D points$\{(x_{i1},y_{i1},z_{i1}), i=\{1,2,..,len(L1)\}\}$, lane boundary $L2$ with 3D points$\{(x_{j2},y_{j2},z_{j2}), j=\{1,2,..,len(L2)\}\}$, constant window size $\eta$, constant threshold $\theta_{dist}$}
\ENSURE{$dict\{(i:j)\}$ as a dictionary of indices for all matched pairs between lane boundary $L1$ and $L2$}
\STATE $N1 \leftarrow len(L1)$
\STATE $N2 \leftarrow len(L2)$
if $N1>N2$, swap$(L1,L2)$ // to ensure $L1$ is not larger than $L2$, i.e. we can always find a matched point in $L2$ with the corresponding point in $L1$ as a pair; Always start searching point pairs from the shorter lane boundary
\STATE $mid1 \leftarrow \frac{N1}{2}$ // the index of the middle number in $L1$, as the start point for searching
\STATE $mid2 \leftarrow index(Y_{ref}^{mid1})$ // generate the search start point in $L2$ from the identical y-reference of $mid1$ 
\STATE $mid2 \leftarrow [\frac{mid2-\eta}{2},\frac{mid2+\eta}{2}]$ and $argmin \underset{mid2}{Dist_{3D}}(L1[mid1],L2[mid2])$ // to find the pair $(mid1, mid2)$ between two lane boundaries

// search backward:
\FOR{($i=mid1-1; i>1; i--$)}
    \STATE $j \leftarrow [mid2-1, mid2-\eta]$ and $argmin \underset{j}{Dist_{3D}}(L1[i],L2[j])$
    \STATE $dict \leftarrow (i,j)$ // to find pair$(i,j)$ and add it to $dict$
    \IF{$|min{Dist_{3D}}(L1[i],L2[j])-min{Dist_{3D}}(L1[i-1],L2[j-1])| > \theta_{dist}$}
        \RETURN NULL // to ensure the difference of lane width is not large locally
    \ENDIF
\ENDFOR

// search forward:
\FOR{($i=mid1+1; i<N1; i++$)}
    \STATE $j \leftarrow [mid2+1, mid2+\eta]$ and $argmin \underset{j}{Dist_{3D}}(L1[i],L2[j])$
    \STATE $dict \leftarrow (i,j)$ // to find pair$(i,j)$ and add it to $dict$
    \IF{$|min{Dist_{3D}}(L1[i],L2[j])-min{Dist_{3D}}(L1[i+1],L2[j+1])| > \theta_{dist}$}
        \RETURN NULL // to ensure the difference of lane width is not large locally
    \ENDIF
\ENDFOR
\RETURN $dict$
\end{algorithmic}
\end{algorithm}

\section{Derivation of the 2D Geometry Constraint}
\label{sec:supequation}
In this section, we derive the 2D geometry prior constraint proposed in main paper. 
First we define the 3D Euclidean distance $D_{3D}$ and 2D Euclidean distance $D_{flat}$ in Equation~\ref{eq:dist_define},

\begin{equation}
\label{eq:dist_define}
\begin{split}
&D_{3D}(P_{i}^{3D},P_{j}^{3D})=|\overrightharpoonup{{P_{i}^{3D}P_{j}^{3D}}}|\\
&=\sqrt{(x_{i}^{3D}-x_{j}^{3D})^2+(y_{i}^{3D}-y_{j}^{3D})^2+(z_{i}^{3D}-z_{j}^{3D})^2}\\
&D_{flat}(P_{i}^{2D},P_{j}^{2D})=|\overrightharpoonup{{P_{i}^{2D}P_{j}^{2D}}}|\\
&=\sqrt{(x_{i}^{2D}-x_{j}^{2D})^2+(y_{i}^{2D}-y_{j}^{2D})^2}\\
\end{split}
\end{equation}

\begin{figure}[ht]
    \centering
    \includegraphics[width=0.8\linewidth]{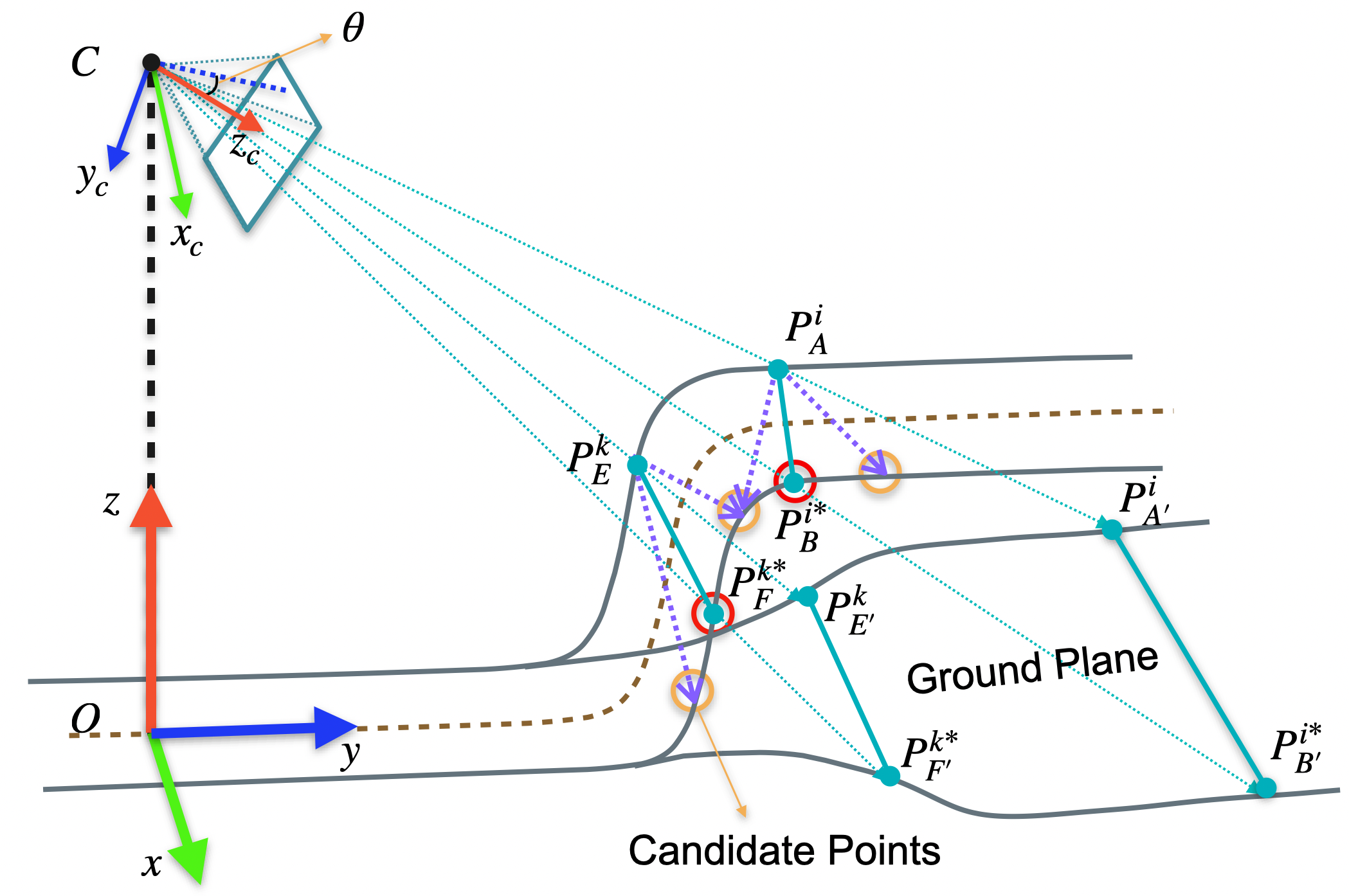}
    \caption{\textbf{Geometry prior of 3D lanes.}}
    \label{fig:lane_width}
\end{figure}

As illustrated in Figure~\ref{fig:lane_width}, we define point $P_{A}^{3D}$ and $P_{E}^{3D}$ in 3D space, and derive the corresponding target points $P_{B}^{3D}$ and $P_{F}^{3D}$ from the formula of parallel curves in 2D parametric representation under the assumption of equal height in the z-axis between corresponding point pairs, as shown in Equation~\ref{eq:predefine}. Let $c$ represent the constant distance between two parallel curves,

\begin{equation}
\label{eq:predefine}
\begin{split}
&P_A^{3D}=\begin{pmatrix}
x_A \\
y_A \\
z_A
\end{pmatrix};
P_B^{3D}=
\begin{pmatrix}
x_A\pm \frac{c}{\sqrt{x_A'^2+y_A'^2}}\cdot y_A' \\
y_A\pm \frac{c}{\sqrt{x_A'^2+y_A'^2}}\cdot x_A'\\
z_A
\end{pmatrix}\\
&P_E^{3D}=
\begin{pmatrix}
x_E \\
y_E \\
z_E
\end{pmatrix};
P_F^{3D}=
\begin{pmatrix}
x_E\pm \frac{c}{\sqrt{x_E'^2+y_E'^2}}\cdot y_E' \\
y_E\pm \frac{c}{\sqrt{x_E'^2+y_E'^2}}\cdot x_E'\\
z_E
\end{pmatrix}\\
\end{split}
\end{equation}
According to the virtual BEV projection~\cite{guo2020gen} in Equation~\ref{eq:virtual_top}, the 3D point is projected onto the ground plane $g$ w.r.t. the camera height $h$ and lane point height $z$. Thus we have the projection of 3D points $P_{A}^{3D}$ and $P_{B}^{3D}$ to the ground as $P_{A}^{2D}$ and $P_{B}^{2D}$ in Equation~\ref{eq:2D_point}.
\begin{equation}
\begin{split}
&P_{}^{2D}=\overrightarrow{C P_{}^{3D}}\cap \Pi_{g}\\
&\begin{pmatrix}
x_{}^{2D} \\
y_{}^{2D} 
\end{pmatrix}
=\frac{h}{h-z} \cdot
\begin{pmatrix}
x_{}^{3D} \\
y_{}^{3D}
\end{pmatrix}
\label{eq:virtual_top}
\end{split}
\end{equation}

\begin{equation}
\begin{split}
&P_{A}^{2D}=
\begin{pmatrix}
x_A\cdot \frac{h}{h-z_A} \\
y_A\cdot \frac{h}{h-z_A} \\
\end{pmatrix} \\
&P_{B}^{2D}=
\begin{pmatrix}
(x_A\pm \frac{c}{\sqrt{x_A'^2+y_A'^2}}\cdot y_A')\cdot \frac{h}{h-z_A} \\
(y_A\pm \frac{c}{\sqrt{x_A'^2+y_A'^2}}\cdot x_A')\cdot \frac{h}{h-z_A} \\
\end{pmatrix}\\
\label{eq:2D_point}
\end{split}
\end{equation}

Following the 3D geometry constraint presented in the paper, we assume the 3D lane has constant width $c$ for each lane point pairs, as described in Equation~\ref{eq:constwidth},
\begin{equation}
\label{eq:constwidth}
\begin{split}
&D_{3D}(P_{A}^{3D},P_{B}^{3D})=D_{3D}(P_{E}^{3D},P_{F}^{3D})=c\\
\end{split}
\end{equation}

thus we have the 2D Euclidean distance of point pairs $\{P_{A}^{2D}, P_{B}^{2D}\}$ and $\{P_{E}^{2D}, P_{F}^{2D}\}$as 
\begin{equation}
\begin{split}
&D_{flat}(P_{A}^{2D},P_{B}^{2D})\\
&=\sqrt{(\frac{h}{h-z_A})^2(\pm \frac{c\cdot y_A'}{\sqrt{x_A'^2+y_A'^2}})^2+(\frac{h}{h-z_A})^2(\pm \frac{c\cdot x_A'}{\sqrt{x_A'^2+y_A'^2}})^2}\\
&=\sqrt{(\frac{h}{h-z_A})^2\cdot (\frac{(c\cdot y_A')^2+(c\cdot x_A')^2)}{x_A'^2+y_A'^2})}\\
&=\sqrt{(\frac{h}{h-z_A})^2\cdot (\frac{(c^2\cdot (y_A'^2+x_A'^2))}{x_A'^2+y_A'^2})}\\
&=\sqrt{(\frac{h}{h-z_A})^2\cdot c^2}\\
&=\boxed{\frac{h}{h-z_A}\cdot c} \text{ $\iff h>z_A$}\\ 
&\text{Similarly, }D_{flat}(P_{E}^{2D},P_{F}^{2D})=\boxed{\frac{h}{h-z_E}\cdot c}\\
\end{split}
\end{equation}
therefore 
\begin{equation}
\begin{split}
D_{flat}(P_{E}^{2D},P_{F}^{2D})\cdot (h-z_E)\\ 
&=h\cdot c\\
&=h\cdot D_{3D}(P_{E}^{3D},P_{F}^{3D})\\
&=h\cdot D_{3D}(P_{A}^{3D},P_{B}^{3D})\\
&=D_{flat}(P_{A}^{2D},P_{B}^{2D})\cdot (h-z_A)\\
\end{split}
\end{equation}
Thus,we define the $D_{2D}$ as the 2D Euclidean distance $D_{flat}$ weighted by camera height $h$ and the same lane height $z$. Thus
\begin{equation}
\begin{split}
D_{2D}(P_{E}^{2D},P_{F}^{2D}) \coloneqq D_{flat}(P_{E}^{2D},P_{F}^{2D}) \cdot (h-z_E) \\
D_{2D}(P_{A}^{2D},P_{B}^{2D}) \coloneqq D_{flat}(P_{A}^{2D},P_{B}^{2D}) \cdot (h-z_A) \\
\end{split}
\end{equation}
therefore under the assumption of shared lane height between points, we have  
\begin{equation}
\begin{split}
D_{2D}(P_{E}^{2D},P_{F}^{2D}) = D_{2D}(P_{A}^{2D},P_{B}^{2D}) \\ 
\label{eq:2D_derive}
\end{split}
\end{equation}

\section{Transformation matrix of augmentation} 
We show the transformation matrix for imposing lane augmentation on pitch (x), roll (y) and yaw (z) axes respectively as
\label{sec:supequationaug}
\begin{equation}
\textbf{R}_x = 
\begin{bmatrix}
1 & 0 & 0 \\
0 & cos \phi & -sin \phi \\
0 & sin \phi & cos \phi
\end{bmatrix} 
\end{equation}
\begin{equation}
\textbf{R}_y = 
\begin{bmatrix}
cos \psi & 0 & sin \psi \\
0 & 1 & 0 \\
-sin \psi & 0 & cos \psi
\end{bmatrix} 
\end{equation}
\begin{equation}
\textbf{R}_z = 
\begin{bmatrix}
cos \theta & -sin \theta & 0 \\
sin \theta & cos \theta & 0 \\
0 & 0 & 1
\end{bmatrix}
\end{equation}

\section{Detailed Experimental Results}
\label{sec:supexplaneline}
In this section, we first provide detailed quantitative results of lane lines in Table~\ref{tab:main1} and center lines in Table~\ref{tab:main2} on the data split of balanced scenes, rarely observed, visual variations and extra-long range following the same setting in the main paper. The joint metric results are evaluated on all methods within the identical data split in the same table, thus cross-split and cross-table comparison of the joint metric is meaningless. Table~\ref{tab:ab_center} presents the ablation study results on center lines, similar to the one for lane lines presented in our main paper.

For our proposed framework, the camera pose for view projection can be estimated by an optional pose regression branch jointly trained with BEV mask in the feature extraction network. As a result, we conduct a comparison for the 3D lane detection under predicted camera pose. We also report BEV mask intersection-over-union (IoU) for a better illustration of the effect on joint training of 2D lane mask and camera pose. As shown in Table~\ref{tab:ab_cam_aug}, when using predicted camera pose, the IoU of predicted lane mask drops for roughly 3\%, however, our proposed augmentation can greatly ease the accuracy drop under unstable camera pose prediction and result in only a slightly drop in the accuracy of the 3D lane detection.

Also, we conduct an ablation study of the proposed augmentation method. As shown in Table~\ref{tab:ab_aug_dim}, the removal of any augmentation method will result in an accuracy drop on both segmentation and lane detection, which proves the effectiveness of all proposed augmentations. Specifically, the roll augmentation should be considered significant for the refinement of the segmentation mask by efficiently generating enriched lane patterns, and pitch augmentation is critical for generating new data with greater fluctuation of lane height which will boost the results of the offset on the far side.

\section{Qualitative comparison}
\label{sec:supvis}
\begin{figure}[t]
    \centering
    \includegraphics[width=0.95\linewidth]{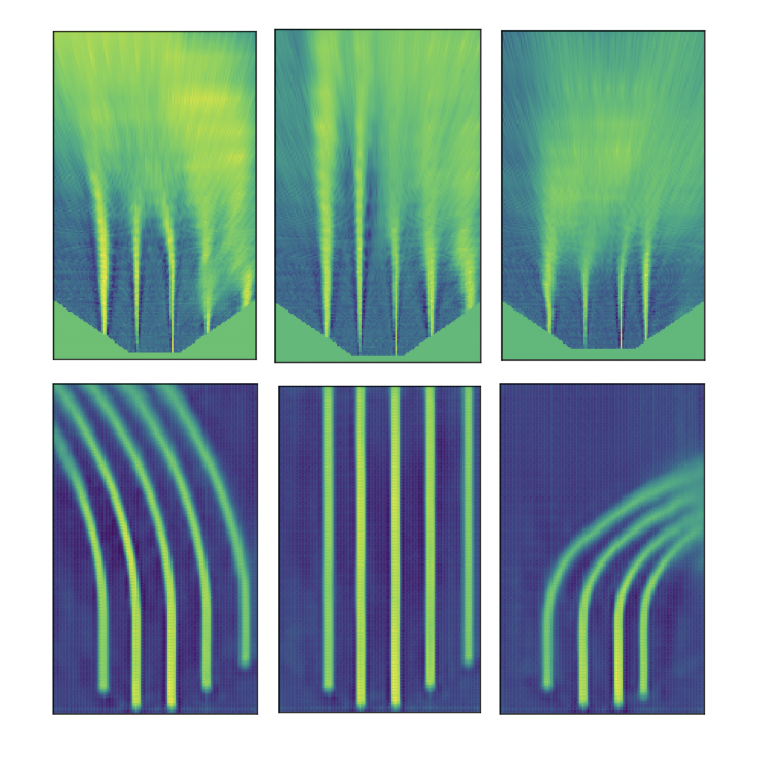}
    \caption{\textbf{Examples of front view supervision with mask projection~\cite{guo2020gen} (top) and our BEV supervision with feature projection (bottom).}}
    \label{fig:bev_extra}
\end{figure}

\begin{figure*}[ht]
    \centering
    \includegraphics[width=0.75\linewidth]{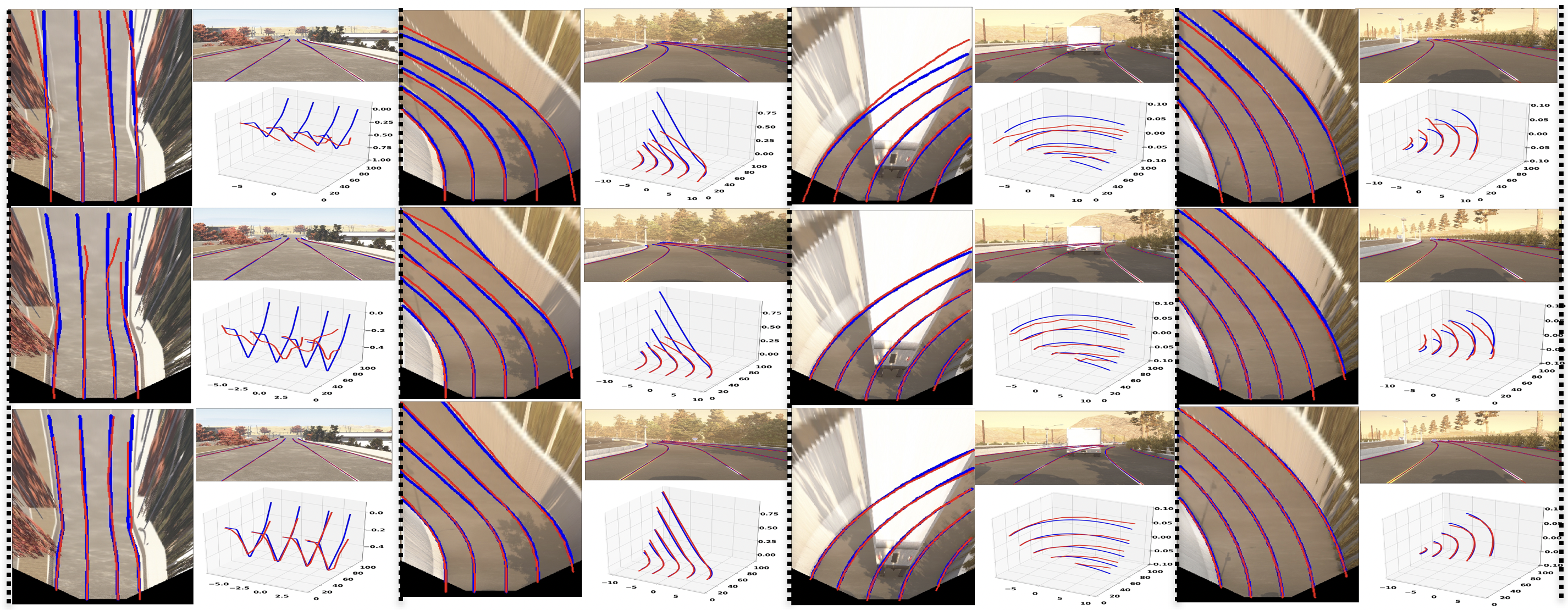}
    \caption{\textbf{Qualitative comparison results of proposed method on lane lines with a maximum range of 100m.} First row: 3D-LaneNet~\cite{garnett20193d}; Second row: Gen-LaneNet~\cite{guo2020gen}; Third row: Our proposed method.}
    \label{fig:lane_vis3}
\vspace{-0.4cm}
\end{figure*}

\begin{figure*}[ht]
    \centering
    \includegraphics[width=0.75\linewidth]{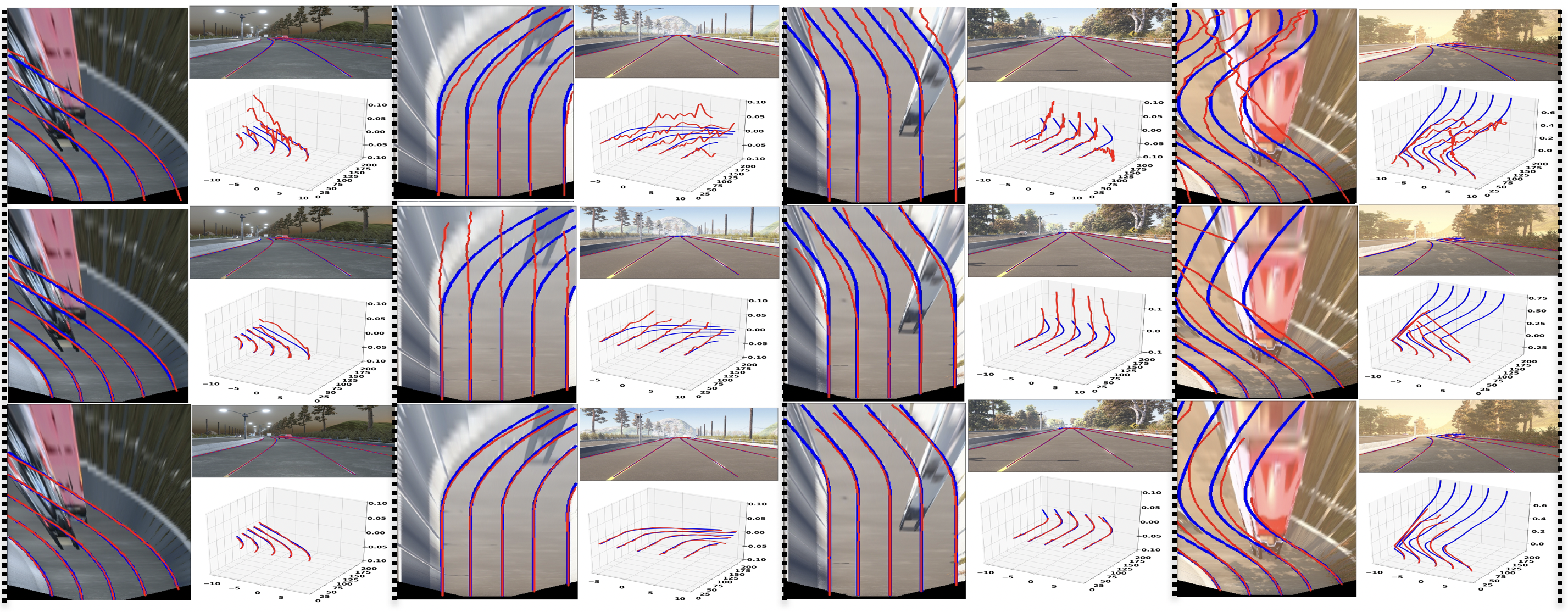}
    \caption{\textbf{Qualitative comparison results of proposed method on lane lines with a maximum range of 200m.} First row: 3D-LaneNet~\cite{garnett20193d}; Second row: Gen-LaneNet~\cite{guo2020gen}; Third row: Our proposed method.}
    \label{fig:lane_vis1}
\vspace{-0.4cm}
\end{figure*}

\begin{figure*}[ht]
    \centering
    \includegraphics[width=0.75\linewidth]{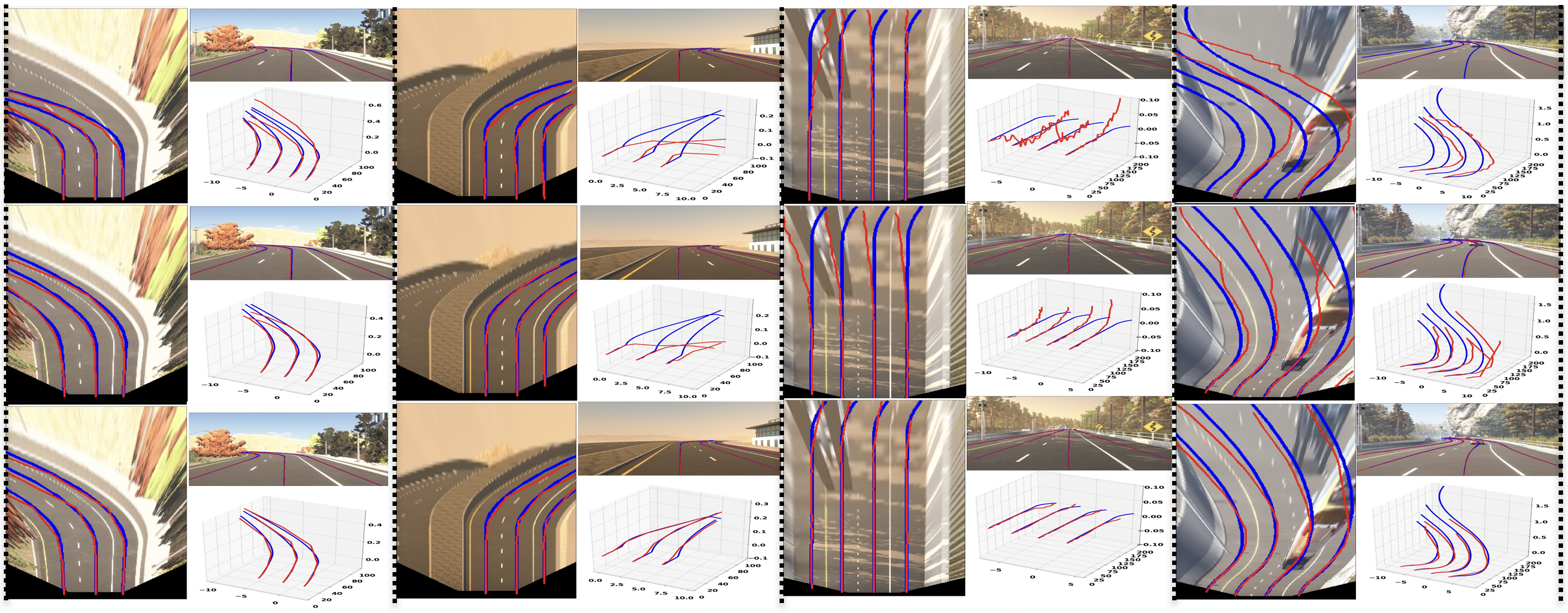}
    \caption{\textbf{Qualitative comparison results of proposed method on center lines.} First row: 3D-LaneNet~\cite{garnett20193d}; Second row: Gen-LaneNet~\cite{guo2020gen}; Third row: Our proposed method.}
    \label{fig:lane_vis2}
\vspace{-0.4cm}
\end{figure*}

We present a detailed qualitative comparison of our proposed method in Figure~\ref{fig:lane_vis3},~\ref{fig:lane_vis1} and~\ref{fig:lane_vis2}. Compared to previous methods, our proposed solution makes tremendously progress in noise elimination, outlier rejection, and structure preservation for 3D lanes. Result shows that our method remains robust even under extreme lightness and strong occlusion. Besides, we provide extra examples on BEV mask predictions of ~\cite{guo2020gen} and ours in \ref{fig:bev_extra}, which proves that our method could tremendously resolve the problem of feature confusion especially in the far end.

\section{Limitation}
\label{sec:suplimit}

\begin{figure}[t]
    \centering
    \includegraphics[width=0.95\linewidth]{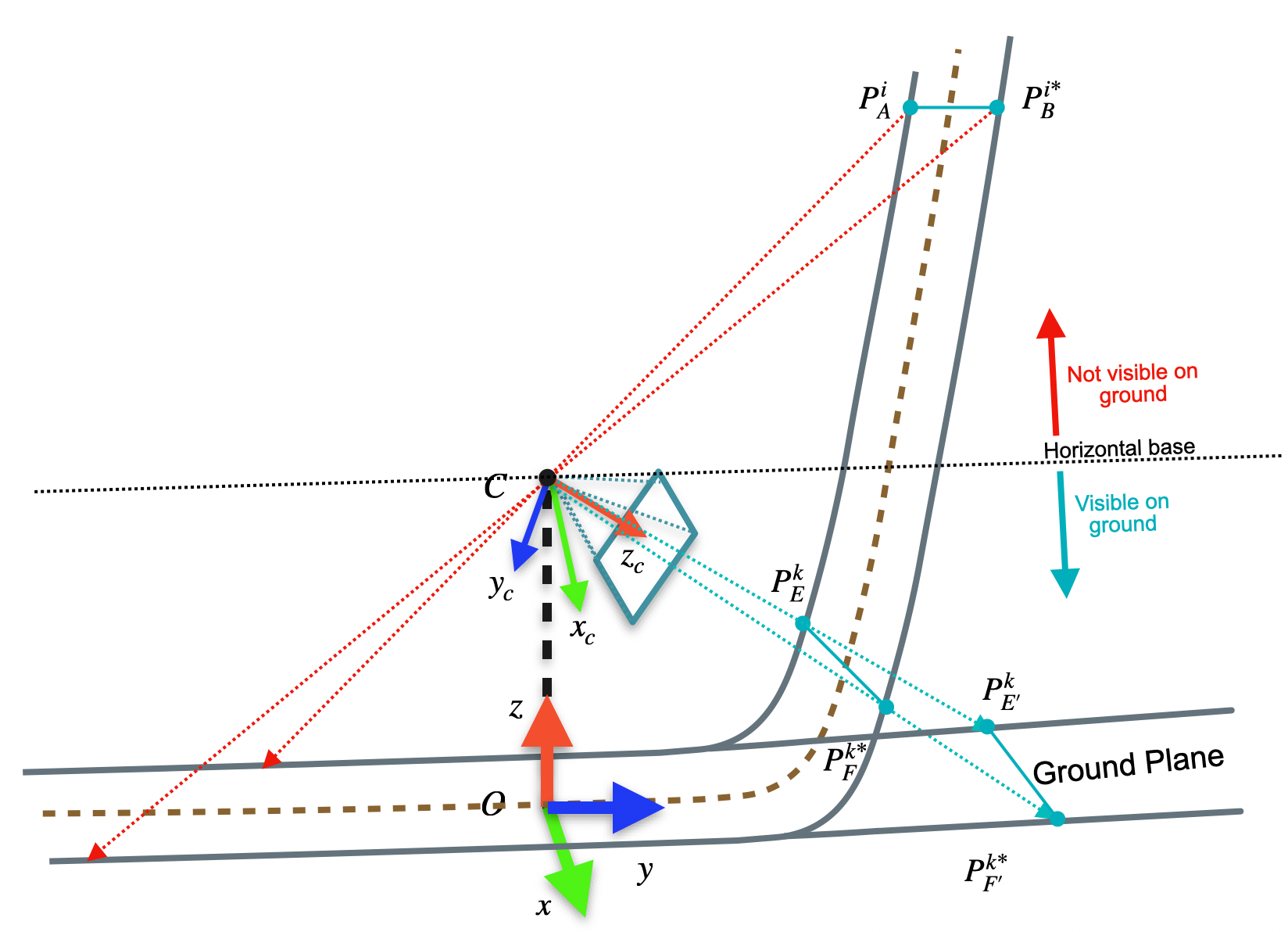}
    \caption{\textbf{An extreme case of lane height}}
    \label{fig:limit}
\end{figure}

In this paper, the proposed pipeline of 3D lane detection is based on 2D-3D lane reconstruction from BEV lane representation. Compared with previous image feature based 3D-LaneNet~\cite{garnett20193d}, the lane-mask based method could tremendously reduce the number of network parameters by extracting the lane height information from the flat ground lane representation under the virtual BEV projection. However, one underlying limitation of such method is the detection range on lane height. For the virtual BEV projection utilized in Gen-LaneNet~\cite{guo2020gen} and our paper, the lane representation on flat ground is generated by projecting 3D lanes via a ray start from the camera center. As a result, as shown in Figure~\ref{fig:limit}, for the cases when part of the lane exceeds the camera height, the 3D lane would be projected to the backside of the camera and become invisible from BEV. In this situation, the network could only make implicit prediction on the out-of-range lanes by following the geometry structure of the visible parts. Even though such hard cases exist, our method could make a certain improvement over Gen-LaneNet~\cite{guo2020gen} on these cases by involving geometry prior.

For the vast majority of lane lines in reality, the lane height would not exceed the camera height within the detection range, which make it a trivial problem in most of the cases. Thus we choose not to propose a detailed solution in this paper for this problem and leave it to the future work.

\section{Future Work}
\label{sec:supfuture}

For the problem mentioned in the limitation section, we consider to address this problem by two solutions in the future work. First, utilizing multi-view feature fusion for a combination of the full lane visibility in front view and the accurate mask representation in BEV. Second, involving the ``virtual camera view'', where multiple virtual cameras are utilized to simulate different installation height for the ensemble of detection results from various view projection. Thus, the edge case of extra-height lanes would be fully considered and the existing pipeline can be more robust under extreme cases.

\begin{table*}
\centering
\caption{Evaluation results on lane lines.}
\label{tab:main1}
\resizebox{\textwidth}{!}{
\begin{tabular}{l|l|cc|cccccc}
\hline
Dataset Split                     & Method               & F-Score & AP   & x error near (m) & x error far (m) & \textbf{joint} x error far (m) & z error near (m) & z error far (m) & \textbf{joint} z error far (m) \\ \hline
\multirow{6}{*}{balanced scenes}   & 3D-LaneNet~\cite{garnett20193d}           & 86.4    & 89.3 & 0.068            & 0.477           &    0.454              & 0.015            & 0.202           &   0.186                    \\
                                   & Gen-LaneNet~\cite{guo2020gen}          & 88.1    & 90.1 & 0.061            & 0.496           &    0.480              & 0.012            & 0.214           &   0.196                 \\
                                   & Ours                 & \textbf{91.9}    & \textbf{93.8} & 0.049            &0.387           &    \textbf{0.340}              & 0.008            & 0.213          &   \textbf{0.175}               \\ \cline{2-10}
                                   & Gen-LaneNet(GT mask)~\cite{guo2020gen}  & 91.8    & 93.8 & 0.054            & 0.412           &    0.361              & 0.011            & 0.226           &   0.180               \\
                                   & Ours (GT mask)       & \textbf{92.8}    & \textbf{94.7} & 0.044           & 0.360          &    \textbf{0.306}              & 0.007            & 0.219          &   \textbf{0.172}               \\ \hline
                                   
\multirow{6}{*}{rarely observed}   & 3D-LaneNet~\cite{garnett20193d}           & 72.0    & 74.6 & 0.166            & 0.855           &    0.906              & 0.039            & 0.521           &   0.551               \\
                                   & Gen-LaneNet~\cite{guo2020gen}           & 78.0    & 79.0 & 0.139            & 0.903           &    0.950              & 0.030            & 0.539           &   0.570                \\
                                   & Ours                 & \textbf{83.7}    & \textbf{85.2} &0.126            & 0.903           &    \textbf{0.870}              & 0.023            & 0.625           &   \textbf{0.567}              \\ \cline{2-10}
                                   & Gen-LaneNet(GT mask)~\cite{guo2020gen}  & 84.7    & 86.6 & 0.117            & 0.839           &    0.785              & 0.024            & 0.611           &   0.548              \\
                                   & Ours (GT mask)       & \textbf{87.8}    & \textbf{89.5} & 0.101           & 0.791           &    \textbf{0.719}              & 0.017            & 0.605           &   \textbf{0.526}               \\ \hline
                                   
\multirow{6}{*}{visual variations} & 3D-LaneNet~\cite{garnett20193d}           & 72.5    & 74.9 & 0.115            & 0.601           &    0.546              & 0.032            & 0.230           &   0.175               \\
                                   & Gen-LaneNet~\cite{guo2020gen}           & 85.3    & 87.2 & 0.074            & 0.538           &    0.444              & 0.015            & 0.232           &   0.161               \\
                                   & Ours                 & \textbf{89.9}    & \textbf{92.1} & 0.06             & 0.446          &   \textbf{ 0.331}              & 0.011            & 0.235          &   \textbf{0.156}             \\ \cline{2-10}
                                   & Gen-LaneNet(GT mask)~\cite{guo2020gen}  & 90.2    & 92.3 & 0.073            & 0.502           &    0.385              & 0.013            & 0.249           &   0.150               \\
                                   & Ours (GT mask)       & \textbf{91.3}    & \textbf{93.2} & 0.055            & 0.435           &    \textbf{0.309}              & 0.010            & 0.249           &   \textbf{0.155}             \\ \hline
                                   
\multirow{6}{*}{extra-long range}  & 3D-LaneNet~\cite{garnett20193d}           & 60.1    & 63.2 & 0.106            & 0.559           &    0.544              & 0.014            & 0.139           &   0.123               \\
                                   & Gen-LaneNet~\cite{guo2020gen}           & 68.5    & 69.2 & 0.064            & 0.524           &    0.503              & 0.010            & 0.112           &   0.088               \\
                                   & Ours                 & \textbf{83.6}    & \textbf{85.3} & 0.039            & 0.290           &    \textbf{0.250}              & 0.007            & 0.087           &   \textbf{0.072}              \\ \cline{2-10}
                                   & Gen-LaneNet(GT mask)~\cite{guo2020gen}  & 80.7    & 82.5 & 0.052            & 0.335           &    0.300              & 0.011            & 0.097           &   0.084               \\
                                   & Ours (GT mask)       & \textbf{87.2}    & \textbf{89.1} & 0.032            & 0.242           &    \textbf{0.221}              & 0.006          & 0.054           &   \textbf{0.047}             \\ \hline
\end{tabular}}
\end{table*}

\begin{table*}[ht]
\centering
\caption{Evaluation results on center lines.}
\label{tab:main2}
\resizebox{\textwidth}{!}{
\begin{tabular}{l|l|cc|cccccc}
\hline
Dataset Split                     & Method                & F-Score & AP   & x error near (m) & x error far (m) & \textbf{joint} x error far (m) & z error near (m) & z error far (m) & \textbf{joint} z error far (m) \\ \hline
\multirow{5}{*}{balanced scenes}   & 3D-LaneNet~\cite{garnett20193d}            & 89.5    & 91.4 & 0.066            & 0.456           & 0.433                 & 0.015            & 0.179           & 0.160                 \\
                                   & Gen-LaneNet~\cite{guo2020gen}            & 90.8    & 92.6 & 0.055            & 0.457           & 0.444                 & 0.011            & 0.176           & 0.167                 \\
                                   & Ours                  & \textbf{94.6}    & \textbf{96.9} & 0.046            & 0.346           & \textbf{0.306}                 & 0.007            & 0.185           & \textbf{0.149}                 \\ \cline{2-10}
                                   & Gen-LaneNet (GT mask)~\cite{guo2020gen}  & 94.5    & 96.8 & 0.050            & 0.372           & 0.325                 & 0.010            & 0.190           & 0.149                 \\
                                   & Ours (GT mask)        & \textbf{95.0}    & \textbf{97.2} & 0.038            & 0.317           & \textbf{0.273}                 & 0.006            & 0.180           & \textbf{0.140}                 \\ \hline
\multirow{5}{*}{rarely observed}   & 3D-LaneNet~\cite{garnett20193d}            & 77.0    & 80.0 & 0.162            & 0.883           & 0.927                 & 0.040            & 0.557           & 0.587                 \\
                                   & Gen-LaneNet~\cite{guo2020gen}            & 79.5    & 80.6 & 0.121            & 0.885           & 0.937                 & 0.026            & 0.547           & 0.606                 \\
                                   & Ours                  & \textbf{84.1}    & \textbf{85.7} & 0.127            & 0.887           & \textbf{0.851 }                & 0.024            & 0.625           & \textbf{0.575}                 \\ \cline{2-10}
                                   & Gen-LaneNet (GT mask)~\cite{guo2020gen}  & 85.9    & 87.7 & 0.105            & 0.845           & 0.812                 & 0.022            & 0.622           & 0.576                 \\
                                   & Ours (GT mask)        &\textbf{87.7}    & \textbf{89.7} & 0.086            & 0.785           & \textbf{0.743}                 & 0.015            & 0.616           & \textbf{0.559}                 \\ \hline
\multirow{5}{*}{visual variations} & 3D-LaneNet~\cite{garnett20193d}            & 75.5    & 77.7 & 0.120            & 0.636           & 0.578                 & 0.030            & 0.227           & 0.174                 \\
                                   & Gen-LaneNet~\cite{guo2020gen}            & 88.2    & 90.0 & 0.072            & 0.438           & 0.430                 & 0.015            & 0.187           & 0.143                 \\
                                   & Ours                  & \textbf{92.2}    & \textbf{94.3} & 0.055            & 0.411           & \textbf{0.300}                 & 0.010            & 0.213           &\textbf{ 0.131}                 \\ \cline{2-10}
                                   & Gen-LaneNet (GT mask)~\cite{guo2020gen}  & 92.3    & 94.2 & 0.071            & 0.467           & 0.356                 & 0.013            & 0.234           & 0.135                 \\
                                   & Ours (GT mask)        & \textbf{93.7}    &\textbf{ 96.1} & 0.055            & 0.397           & \textbf{0.281}                 & 0.009            & 0.222           & \textbf{0.130}                 \\ \hline
\multirow{5}{*}{extra-long range}  & 3D-LaneNet~\cite{garnett20193d}            & 62.2    & 64.0 & 0.106            & 0.559           & 0.526                 & 0.014            & 0.139           & 0.071                 \\
                                   & Gen-LaneNet~\cite{guo2020gen}            & 69.4    & 70.1 & 0.058            & 0.507           & 0.465                 & 0.009            & 0.082           & 0.040                 \\
                                   & Ours                  & \textbf{85.8}    & \textbf{87.6} & 0.037            & 0.264           &\textbf{0.204}                 & 0.007            & 0.067           & \textbf{0.032}                 \\ \cline{2-10}
                                   & Gen-LaneNet (GT mask)~\cite{guo2020gen}  & 83.0    & 84.3 & 0.050            & 0.313           & 0.251                 & 0.011            & 0.076           & 0.035                 \\
                                   & Ours (GT mask)        & \textbf{88.6}    & \textbf{90.4} & 0.031            & 0.205           & \textbf{0.147}                 & 0.007            & 0.054           & \textbf{0.028}                 \\ \hline
\end{tabular}}
\end{table*}

\begin{table*}[h]
\centering
\caption{Ablation study on center line prediction.}
\label{tab:ab_center}
\resizebox{\textwidth}{!}{
\begin{tabular}{l|cc|cccccc}
\hline
Method                 & F-Score & AP   & x error near (m) & x error far (m) & \textbf{joint} x error far (m) & z error near (m) & z error far (m) & \textbf{joint} z error far (m) \\ \hline
3D-LaneNet~\cite{garnett20193d}             & 89.5    & 91.4 & 0.066            & 0.456           & 0.427                 & 0.015                 & 0.179                      & 0.157                 \\
Gen-LaneNet~\cite{guo2020gen}             & 90.8    & 92.6 & 0.055            & 0.457           & 0.435                & 0.011                 & 0.176            & 0.164                 \\
Ours w/ GS             & 92.3    & 94.2 & 0.047            & 0.415           & 0.372               & 0.008                 & 0.186               & 0.150                 \\
Ours w/ BVS            & 92.7    & 94.7 & 0.059            & 0.411           & 0.360               & 0.012                 & 0.208               & 0.158                 \\
Ours w/ GS + BVS       & 93.7    & 95.9 & 0.062            & 0.377           & 0.333              & 0.008                 & 0.187               & 0.150                 \\
Ours w/ GS + BVS + Aug & \textbf{94.6}    & \textbf{96.9} & 0.046            & 0.346           & \textbf{0.300}              & 0.007                 & 0.185               & \textbf{0.146}                 \\ \hline
Gen-LaneNet (GT mask)~\cite{guo2020gen}   & 94.5    & 96.8 & 0.050            & 0.372           & 0.318            & 0.010                 & 0.190                  & 0.145                 \\
Ours (GT mask)         &\textbf{ 95.0}    & \textbf{97.2} & 0.038            & 0.317           & \textbf{0.268}            & 0.006                 & 0.180                 & \textbf{0.137}                 \\ \hline
\end{tabular}}
\end{table*}

\begin{table*}[h]
\centering
\caption{Ablation study on camera pose and augmentation.}
\label{tab:ab_cam_aug}
\resizebox{\textwidth}{!}{
\begin{tabular}{c|l|c|cc|cccccc}
\hline
Task                                & Method                & BEV mask IoU & F-Score & AP   & x error near (m) & x error far (m) & \textbf{joint} x error far (m) & z error near (m) & z error far (m) & \textbf{joint} z error far (m) \\ \hline
\multirow{4}{*}{Lane line}          & Pred cam pose w/o Aug & 91.7              & 90.1    & 92.2 & 0.064            & 0.447           & 0.426                 & 0.017            & 0.243           &   0.222                    \\
                                    & Pred cam pose w Aug   & 92.9              & 91.0    & 93.2 & 0.066            & 0.444           & 0.407                 & 0.019            & 0.240           &   0.222               \\
                                    & GT cam pose w/o Aug   & 94.7              & 91.2    & 93.2 & 0.065            & 0.415           & 0.394                 & 0.009            & 0.220           &   0.207               \\
                                    & GT cam pose w Aug     & \textbf{96.4}              & \textbf{91.9}    &\textbf{ 93.8} & 0.049            & 0.387           & \textbf{0.363}                 & 0.008            & 0.213           &   \textbf{0.200}               \\ \hline
\multirow{4}{*}{Center line}        & Pred cam pose w/o Aug & 91.7              & 92.9    & 94.9 & 0.058            & 0.419           & 0.394                 & 0.017            & 0.213           &   0.196               \\
                                    & Pred cam pose w Aug   & 92.9              & 93.3    & 95.4 & 0.063            & 0.408           & 0.369                 & 0.018            & 0.212           &   0.194               \\
                                    & GT cam pose w/o Aug   & 94.7              & 93.7    & 95.9 & 0.062            & 0.377           & 0.354                 & 0.008            & 0.187           &   0.173                \\
                                    & GT cam pose w Aug     & \textbf{96.4}              & \textbf{94.6}    & \textbf{96.9} & 0.046            & 0.346           & \textbf{0.322}                 & 0.007            & 0.185           &   \textbf{0.169}               \\ \hline
\end{tabular}}
\end{table*}

\begin{table*}[h]
\centering
\caption{Ablation study on 3D lane augmentation on different axes.}
\label{tab:ab_aug_dim}
\resizebox{0.9\textwidth}{!}{
\begin{tabular}{c|l|c|cc|cccc}
\hline
Task                                & Method                          & BEV mask IoU & F-Score & AP   & x error near (m) & x error far (m) & z error near (m) & z error far (m) \\ \hline
\multirow{4}{*}{Lane line}          & No Aug                          & 94.7              & 91.2    & 93.2 & 0.065            & 0.415                                  & 0.009            & 0.220                                  \\
                                    & Pitch + Yaw (w/o Roll)          & 95.0              & 91.3    & 93.3 & 0.051            & 0.402                                  & 0.012            & 0.219                                  \\
                                    & Pitch + Roll (w/o Yaw)          & 95.4              & 91.4    & 93.4 & 0.049            & 0.414                                  & 0.010            & 0.217                                  \\
                                    & Yaw + Roll (w/o Pitch)          & 95.4              & 91.2    & 93.3 & 0.060            & 0.438                                  & 0.011            & 0.228                                  \\
                                    & Pitch + Yaw + Roll (Ours)       & \textbf{96.4}              & \textbf{91.9}    & \textbf{93.8} & 0.049            & 0.387                                  & 0.008            & 0.213                                  \\ \hline
\multirow{4}{*}{Center line}        & No Aug                          & 94.7              & 93.7    & 95.9 & 0.062            & 0.377                                  & 0.008            & 0.187                                  \\
                                    & Pitch + Yaw (w/o Roll)          & 95.0              & 94.2    & 96.6 & 0.050            & 0.359                                  & 0.010            & 0.188                                  \\
                                    & Pitch + Roll (w/o Yaw)          & 95.4              & 93.7    & 95.8 & 0.047            & 0.365                                  & 0.008            & 0.187                                  \\
                                    & Yaw + Roll (w/o Pitch)          & 95.4              & 93.5    & 95.7 & 0.057            & 0.386                                  & 0.010           & 0.191                                 \\
                                    & Pitch + Yaw + Roll (Ours)       & \textbf{96.4}              & \textbf{94.6}    & \textbf{96.9} & 0.046            & 0.346                                  & 0.007            & 0.185                                  \\ \hline
\end{tabular}}
\end{table*}

\newpage

{\small
\bibliographystyle{ieee_fullname}
\bibliography{egbib}
}